%% file: paper_neuromorphic_dvfs.tex
\documentclass[10pt,journal]{IEEEtran}

\usepackage[utf8]{inputenc}
\usepackage{amsmath}
\usepackage{subfigure}
\usepackage{cite}
\usepackage{color}
\usepackage{siunitx}
\usepackage{booktabs}
\usepackage[table]{xcolor}
\usepackage{colortbl}
\usepackage{multirow}

\definecolor{tablecol-even}{gray}{0.97}
\definecolor{tablecol-odd}{gray}{1.0}
\newcommand{\setrowcolors}{\rowcolors{2}{tablecol-even}{tablecol-odd}}
\newcommand{\unsetrowcolors}{\rowcolors{0}{}{}}

\usepackage[noreledmac]{eledmac}

\foottwocolX{A}
\newcommand\tablenote[1]{\footnoteA{#1}}

\usepackage{ifpdf}
\ifpdf
  \usepackage[pdftex]{graphicx}
  \else \usepackage[dvips]{graphicx}
\fi

\renewcommand{\t}[1]{
\textnormal{#1}
}

\input{authorblock_ieee.tex}

\begin{document}
© 2019 IEEE.  Personal use of this material is permitted.  Permission from IEEE must be obtained for all other uses, in any current or future media, including reprinting/republishing this material for advertising or promotional purposes, creating new collective works, for resale or redistribution to servers or lists, or reuse of any copyrighted component of this work in other works.
\clearpage
\bstctlcite{IEEEexample:BSTcontrol}
\maketitle

\begin{abstract}
\input{abstract.tex}

\end{abstract}

\begin{IEEEkeywords}
MPSoC, neuromorphic computing, SpiNNaker2, power management, DVFS, synfire chain
\end{IEEEkeywords}

\section{Introduction}
\input{introduction.tex}

\section{Neuromorphic SoC Architecture}
\label{sec:neurosocarchitecture}

\subsection{Overview}
\input{many_core_system_architecture.tex}

\subsection{Power Management Hardware Architecture}
\label{sec:powermanagementhardwarearchiecture}
\input{power_management_architecture.tex}

\subsection{Spiking Neural Network Simulation}
\input{spiking_neural_network_simulation.tex}

\subsection{Power Management Software Architecture}
\input{power_management_software.tex}
\section{Power and Energy Model}
\label{sec:pmtheory}
\input{pm_theory.tex}

\section{Test Chip}
\label{sec:testchipsantos}
\input{testchip_santos.tex}

\section{Results}
\label{sec:results}
\input{results.tex}

\section{DVFS Architecture Exploration}
\label{sec:dvfs_architecture_explorations}
\input{dvfs_architecture_exploration.tex}

\section{Conclusion}
\input{conclusion.tex}
\label{sec:conclusion}


\bibliographystyle{IEEEtran}
\inputencoding{latin1}
\bibliography{bib_library}
\inputencoding{utf8}

\input{bio.tex}

\end{document}

%% file: authorblock_ieee.tex
\title{Dynamic Power Management for Neuromorphic Many-Core Systems}

\author{Sebastian Höppner, Bernhard Vogginger, Yexin Yan, Andreas Dixius, Stefan Scholze, Johannes Partzsch, Felix Neumärker, Stephan Hartmann, Stefan Schiefer, Georg Ellguth,  Love Cederstroem, Luis Plana, Jim Garside, Steve Furber, Christian Mayr
\thanks{Manuscript received \today. The research leading to these results has received funding from the European Union Seventh Framework Programme (FP7) under grant agreement no 604102 and the EUs Horizon 2020 research and innovation programme under grant agreements No 720270 and 785907 (Human Brain Project, HBP) and the Center for Advancing Electronics Dresden (cfaed). The authors thank ARM and Synopsis for IP and the Vodafone Chair at Technische Universität Dresden for contributions to RTL design.}
\thanks{S. Höppner, B. Vogginger, Y. Yan, A. Dixius, S. Scholze, J. Partzsch, F. Neumärker, S. Hartmann, S. Schiefer, G. Ellguth, L. Cederstroem and C. Mayr are with the Faculty of Electrical and Computer Engineering, Technische Universität Dresden, Germany (e-mail: sebastian.hoeppner@tu-dresden.de) }
\thanks{L. Plana, J. Garside and S. Furber are with the Advanced Processor Technologies Research Group, University of Manchester}
}

%% file: abstract.tex
This work presents a dynamic power management architecture for neuromorphic many core systems such as SpiNNaker.
A fast dynamic voltage and frequency scaling (DVFS) technique is presented which allows the processing elements (PE) to change their supply voltage and clock frequency individually and autonomously within less than \SI{100}{ns}.
This is employed by the neuromorphic simulation software flow, which defines the performance level (PL) of the PE based on the actual workload within each simulation cycle.
A test chip in \SI{28}{nm}~SLP CMOS technology has been implemented.
It includes \SI{4}{PEs} which can be scaled from \SI{0.7}{V} to \SI{1.0}{V} with frequencies from \SI{125}{MHz} to \SI{500}{MHz} at three distinct PLs.
By measurement of three neuromorphic benchmarks it is shown that the total PE power consumption can be reduced by \SI{75}{\%}, with \SI{80}{\%} baseline power reduction and a \SI{50}{\%} reduction of energy per neuron and synapse computation, all while maintaining temporary peak system performance to achieve biological real-time operation of the system.
A numerical model of this power management model is derived which allows DVFS architecture exploration for neuromorphics.
The proposed technique is to be used for the second generation SpiNNaker neuromorphic many core system.

%% file: introduction.tex
%
%

Neuromorphic circuits~\cite{Furber2016} try to emulate aspects of neurobiological information in semiconductor hardware in order to solve problems that biology excels at, for example robotics control, image processing or data classification. Furthermore, it represents an alternative to general-purpose high-performance computing for large-scale brain simulation~\cite{vanAlbada2018performance}. Besides system capacity, energy efficiency is one major scaling target, especially when large scale brain models are to be simulated with reasonable power consumption, which is typically limited by the effort for cooling and power supply. Energy efficiency is mandatory for the application in mobile, battery powered scenarios such as drones or mobile robots.

There exist approaches for low power and energy efficient neuromorphic circuits using analog subthreshold circuits~\cite{Yu2012,qiao2015reconfigurable,Indiveri2015,Benjamin2014} or in recent years memristors~\cite{du15,mostafa15}. However, these systems show severe device variability and it is challenging to scale them to larger systems. In contrast, digital neuromorphic hardware systems such as TrueNorth~\cite{Akopyan2015}, Loihi~\cite{Davies2018} or SpiNNaker~\cite{Painkras2013} emulate neural processing by means of digital circuits or embedded software. Due to their purely digital realization they can be implemented in the latest CMOS technologies, operate very reliably/reproducibly and can be scaled to large system sizes.

To make digital neuromorphic systems as energy-efficient as analog ones, we can take inspiration from biology: The brain seems to maximize the ratio of information transmitted/computed to energy consumed~\cite{harris2012}. That is, it consumes energy proportional to task complexity or activity levels. The brain also seems to use this in a spatial dimension, i.e.\ energy is allocated to different regions according to task demand~\cite{lennie2003}. The cost of a single spike limits the number of neurons that are concurrently active to one percent of the brain.

Transferring this concept to digital neuromorphic hardware, where large neural networks are mapped to many processing cores, leads to the requirement for local, fine-grained scalability for the trade-off between processing power and energy efficiency for highly dynamic workloads, i.e.\ spikes to be processed in biological real time. The temporal demand for high computational performance to be able to process large neural networks in biological real time requires high processor clock frequencies, and thereby the operation at nominal or even overdrive supply voltages, leading to high-power consumption. In contrast, low power operation can only be achieved by operating the processor core at lower supply voltages at the cost of higher logic delays and thereby slower maximum clock frequencies. Dynamic voltage and frequency scaling (DVFS) during system operation can break this trade-off, by lowering supply voltage and frequency during periods of low processor performance requirements and increasing supply voltage and clock frequency only if peak performance is required temporarily~\cite{Ma2010}.

DVFS is widely used in modern MPSoCs, like for mobile communication~\cite{Winter2012,Noethen2014,Haas2017} or database acceleration~\cite{Haas2016}. In the mentioned scenarios DVFS is typically controlled by a task scheduling unit which assigns supply voltage and frequency settings to the worker processing elements dynamically. In contrast, neuromorphic many core systems do not contain a central task management unit, since the neuromorphic application runs in parallel on processing elements~\cite{Painkras2013}. Their actual workload for neuron state updates and synaptic event processing mainly depends on both the neural network topology and the input (e.g.\ spikes) of the experiment or application. It is therefore not known in advance or by any central control unit. However, the DVFS technique is beneficial for those systems, since neural networks show significant temporal variations of activity, making them very energy efficient for the neuromorphic task to be solved. This technique is to be employed by future digital neuromorphic hardware, such as the second generation of the SpiNNaker~\cite{Painkras2013} system, which is currently in the prototyping phase.

\textcolor{red}{This work presents a DVFS architecture for neuromorphic many core systems, where each processing element can autonomously change its supply voltage and frequency, only determined by its local workload. It extends a conference publication~\cite{Hoeppner2017} by a technique for workload estimation and performance level selection for neuromorphic DVFS and a numerical energy consumption model for architecture exploration. Both aspects are demonstrated by means of chip measurement results for several benchmarks.} In Sec.~\ref{sec:neurosocarchitecture} the architecture of the neuromorphic SoC is presented including the hardware architecture and software flow for DVFS and Sec.~\ref{sec:pmtheory} describes the corresponding power consumption model. Sec.~\ref{sec:testchipsantos} presents a test chip in 28nm CMOS, with measurement results including 3 neuromorphic benchmarks summarized in Sec.~\ref{sec:results}. Sec.~\ref{sec:dvfs_architecture_explorations} shows an exploration of DVFS architecture parameters for future neuromorphic many core systems, such as SpiNNaker2.


%% file: many_core_system_architecture.tex
%
%

\begin{figure}[htb]
	\centering
		\includegraphics[width=0.47\textwidth]{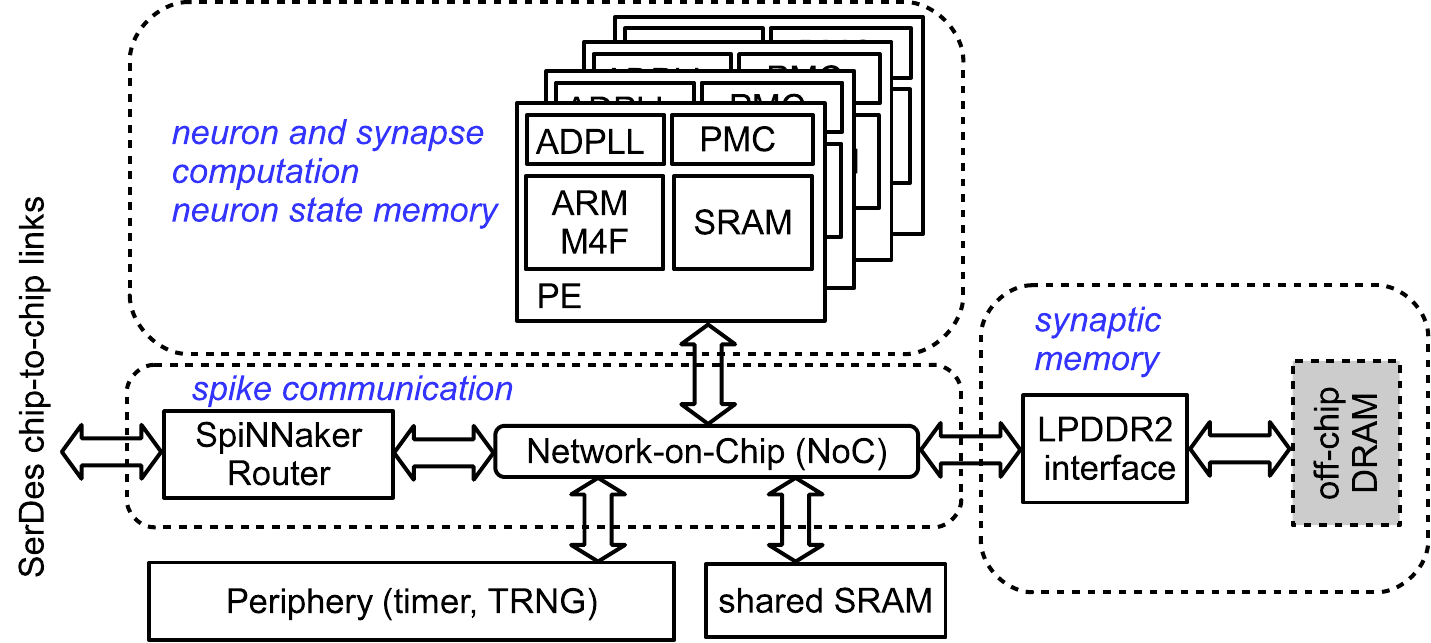}
	\caption{System architecture}\label{fig:system_architecture}
\end{figure}

Fig.\ref{fig:system_architecture} shows the block diagram of the many core system, based on the SpiNNaker architecture~\cite{Painkras2013}. The neuromorphic computation problem is mapped to processing elements (PEs), which are responsible for neuron state computation and synaptic processing. In this work, the PEs are based on ARM M4F processors with local SRAM and a globally asynchronous locally synchronous (GALS) clocking architecture for dynamic power management. Off chip DRAM serves as synaptic weight storage. A SpiNNaker router~\cite{Navaridas201549} is used for on-chip and off-chip neuromorphic spike communication. The on-chip components are connected by a packet based network-on-chip (NoC), carrying spike packets, DMA packets for DRAM access and various types of control packets. In the periphery, shared modules are included, such as true random number generators (TRNGs)~\cite{Neumarker2016} and timers for the neuromorphic real time simulation time step (e.g. 1ms) generation.

%% file: power_management_architecture.tex
\begin{figure}[htb]
	\centering
		\includegraphics[width=0.47\textwidth]{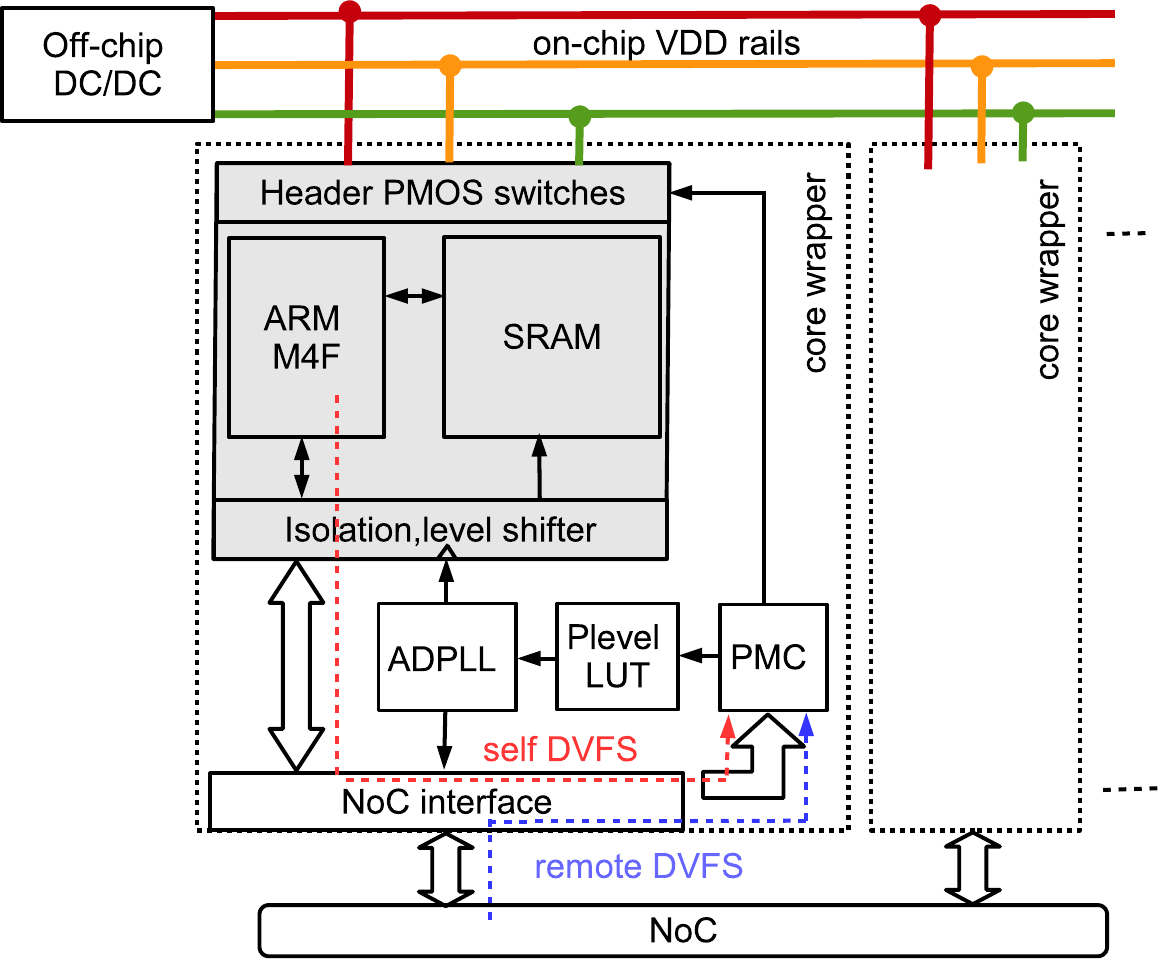}
	\caption{PE DVFS architecture~\cite{Hoeppner2017}}
	\label{fig:PE_dvfs_architecture}
\end{figure}

%
%
The power management architecture of the PE is shown in Fig.~\ref{fig:PE_dvfs_architecture}. It is based on the concepts from~\cite{Hoeppner2012b} and~\cite{Noethen2014}. The PE is equipped with a local \textcolor{red}{all-digital phase-locked-loop} ADPLL~\cite{Hoeppner2013} with open-loop output clock generation~\cite{Hoeppner2011a}, enabling ultra-fast defined frequency switching. The core domain, including the processor and local SRAM is connected to one out of several global supply rails by PMOS header power switches. If all switches are opened, the core is in power-shut-off. This allows for individual, fine-grained power management and GALS operation.

\begin{figure}[htb]
	\centering
		\includegraphics[width=0.47\textwidth]{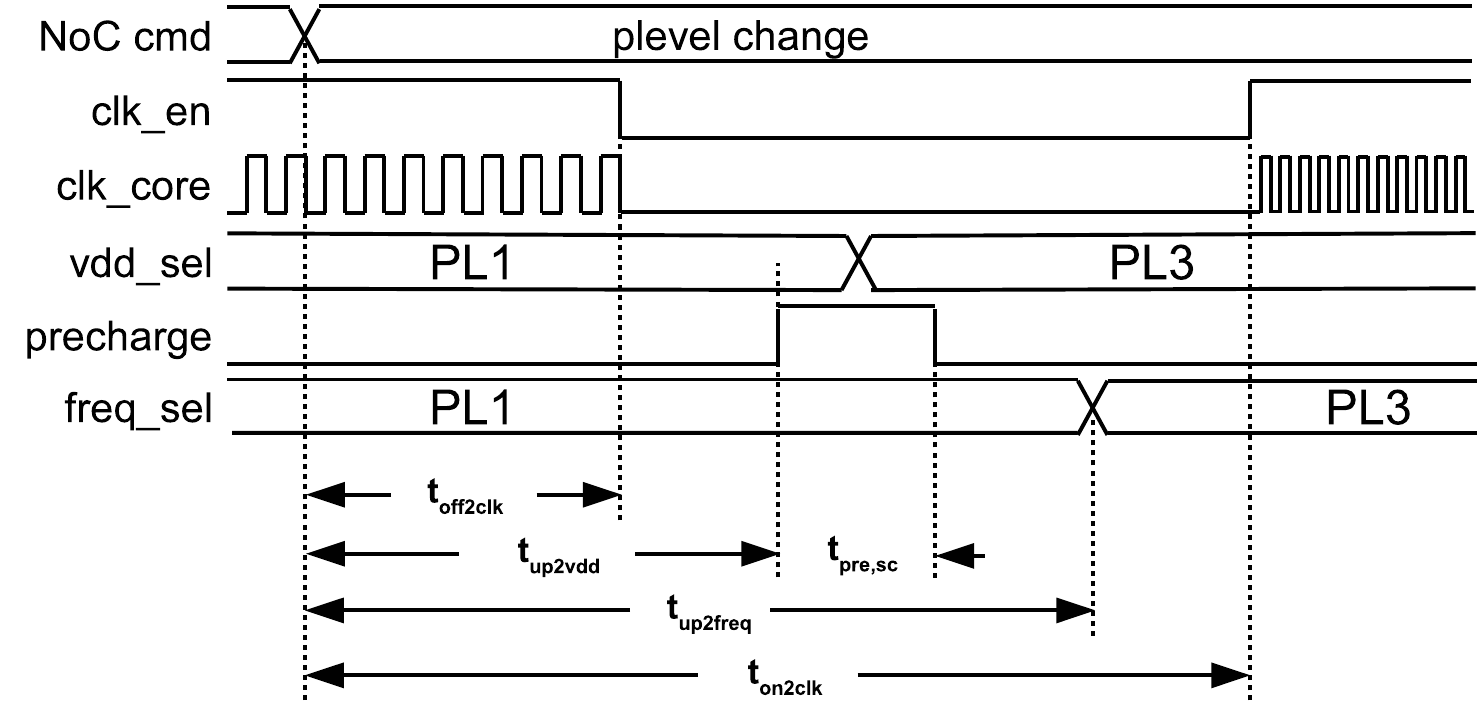}
	\caption{PE DVFS Timing of performance level change~\cite{Hoeppner2017}}
	\label{fig:PE_dvfs_timing}
\end{figure}

A PE performance level (PL) consists of a supply voltage and frequency $(V_\t{DD},f)$ pair. Switching between PLs is scheduled by the power management controller (PMC)~\cite{Hoeppner2012b}. The supported scenarios include the change between two PLs when the core is active as well as power-shut-off (PSO) and power-on after PSO. Therefore, the PMC schedules a sequence of the events clock disable, supply selection, net pre-charge, frequency selection and clock enable as shown in Fig.~\ref{fig:PE_dvfs_timing}. The timings are derived from the reference clock signal (period typically \SI{10}{ns}) and are fully configurable.

\begin{figure}[htb]
	\centering
		\includegraphics[width=0.30\textwidth]{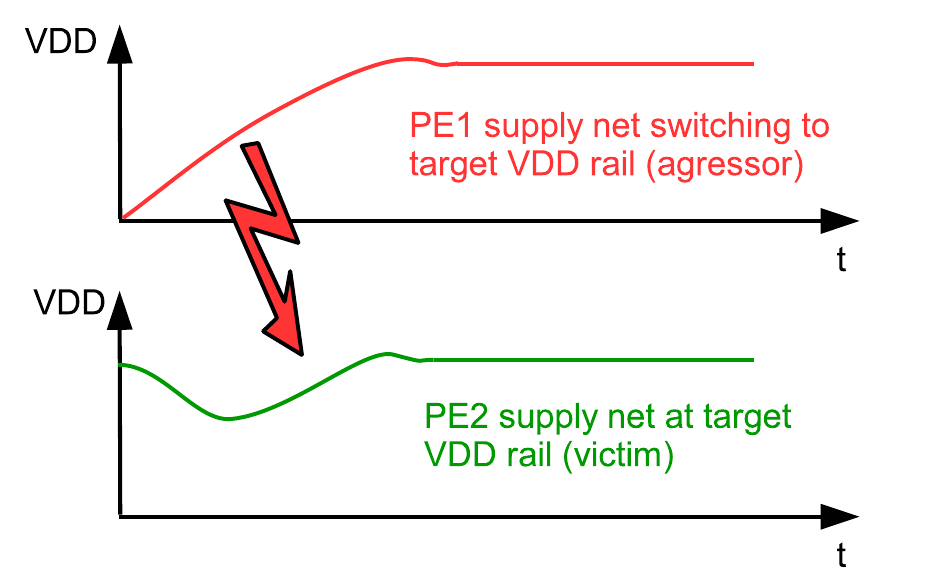}
	\caption{Illustration of PE supply switching}
	\label{fig:PE_dvfs_supply_xtalk}
\end{figure}

When connecting a PE to another supply rail at a new target $V_\t{DD}$ level, rush currents must be reduced to prevent unwanted supply voltage drops of other PE, being currently operated at the target rail, as illustrated in Fig.~\ref{fig:PE_dvfs_supply_xtalk}. Therefore, a pre-charge scheme is used, where a small (configurable) number of power switches is connected to the target supply net to reduce the slew rate of the switched net~\cite{Hoeppner2012b}. Measurement results of this scheme for the \SI{28}{nm} SoC implementation in this work are shown in Sec.~\ref{sec:pemeasresults}. As result PL changes of active PEs can be achieved within approximately \SI{100}{ns}, i.e., \emph{instantaneously} from the perspective of the software running on the PE.

PL changes are triggered by sending commands via the NoC interface to the PMC. For power-up or remotely controlled DVFS, these commands can be sent by another core, orchestrating system boot-up. During PE operation (e.g.\ distributed neuromorphic application) the PE can trigger PL changes on its own by sending a NoC packet to its own PMC (\emph{self DVFS}). Thus, the PE software can actively change its PL without significant latency or software overhead. The application specific power management algorithms can be completely implemented in software at the local PEs.

%% file: spiking_neural_network_simulation.tex
%
%

\begin{figure}[htb]
	\centering
	\subfigure[Logical SNN realization \label{fig:SNN_architecture_logical}]{\includegraphics[width=0.4\textwidth]{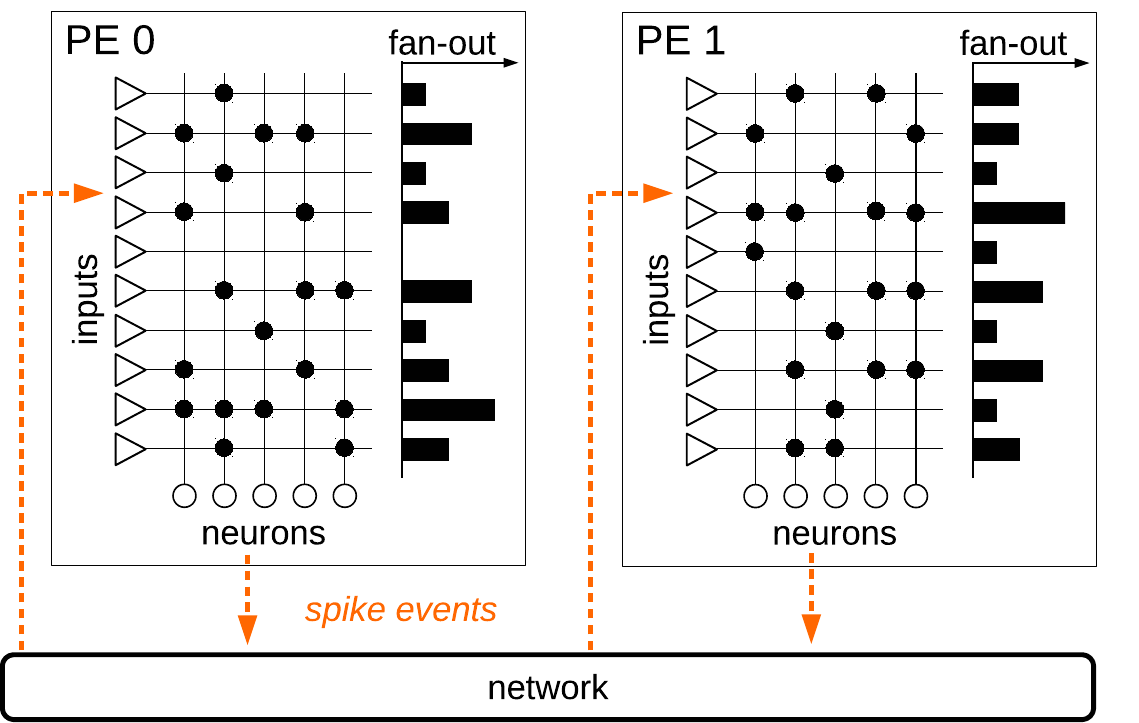}}
	\subfigure[Physical implementation \label{fig:SNN_architecture_impl}]{\includegraphics[width=0.47\textwidth]{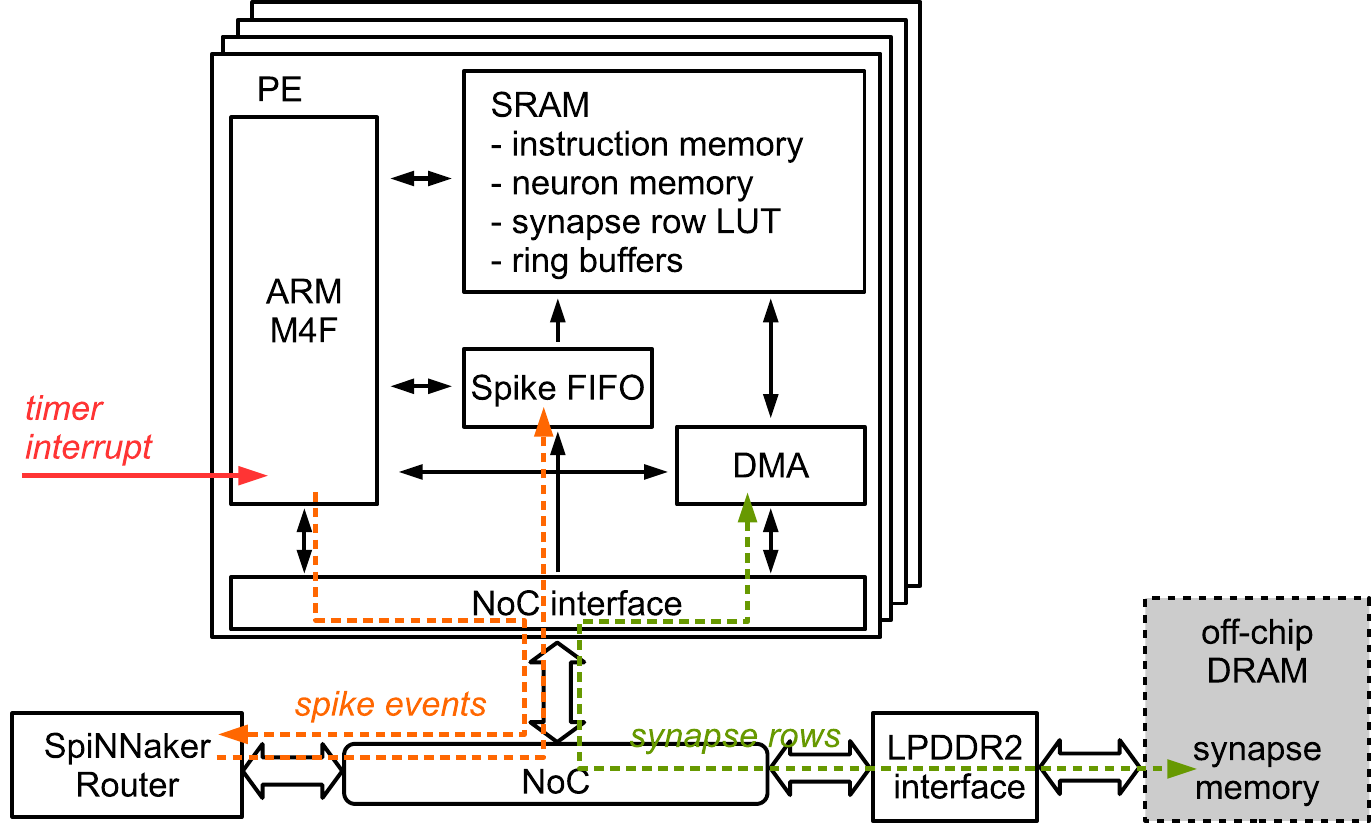}}
  \caption{\textcolor{red}{Spiking neural network simulation architecture}}
	\label{fig:SNN_architecture}
\end{figure}

We follow the approach of SpiNNaker~\cite{Furber2014} to implement real-time \textcolor{red}{spiking neural networks (SNN)}. Each core simulates the dynamics of a number of neurons and their inbound synapses (Fig.~\ref{fig:SNN_architecture_logical}). A real-time tick from a peripheral timer regularly triggers the neuron state updates and synapse processing which must complete within each simulation cycle $t_\t{sys}$ (e.g. \SI{1}{\ms}) otherwise the neurons may fall out of sync. The spike communication between neurons is established by the SpiNNaker router
that forwards multicast packets containing the identifiers of the sending neurons to target PEs according to a configurable routing table. At the target PE the spike events are inserted into a hardware FIFO attached to the local SRAM and processed in the subsequent cycle. Note that this is different to the typical SpiNNaker operation where incoming
spikes are processed immediately.

Details about the memory layout for neural processing are shown in Fig.~\ref{fig:SNN_architecture_impl}. Being accessed in every simulation cycle, the neuron state variables and parameters are stored in local SRAM. The synapse parameters, which require more memory and are only needed at an incoming spike, are stored in external DRAM. They are organized in so-called \emph{synapse rows} which are contiguous memory blocks containing the synapses between one source neuron and all neurons of a core~\cite{noack10}.
In the synapse row each existing synaptic connection is represented by a 32-bit word with a 16-bit weight, an 8-bit target neuron identifier, one
synapse type bit (excitatory/inhibitory), and a 4-bit delay. The size of a synapse row depends on the fan-out of each source neuron, cf.
Fig.~\ref{fig:SNN_architecture_logical}.

When processing a received spike event, the core extracts start address and size of the synapse row belonging to the source neuron from a lookup table in
the SRAM. Then, a direct memory access (DMA) for reading the synapse row from the external DRAM is scheduled with a dedicated DMA controller.
During the DMA transfer the processor is not idle but can execute other tasks like neuron updates from a job queue~\cite{Sharp2011}.
Upon completion of the DMA, the synapses are processed and the weights are added to input ring buffers of the target neurons.
These ring buffers accumulate the synaptic inputs for the next \SI{15}{clock} cycles
and enable configurable synaptic delays.
When calculating the neuron state update, the synaptic input from the buffer corresponding to the current cycle is injected into the neuron model.
If neurons have fired, spike events are generated and sent to the SpiNNaker router.





%% file: power_management_software.tex
\begin{figure}[htb]
	\centering
		\includegraphics[width=0.47\textwidth]{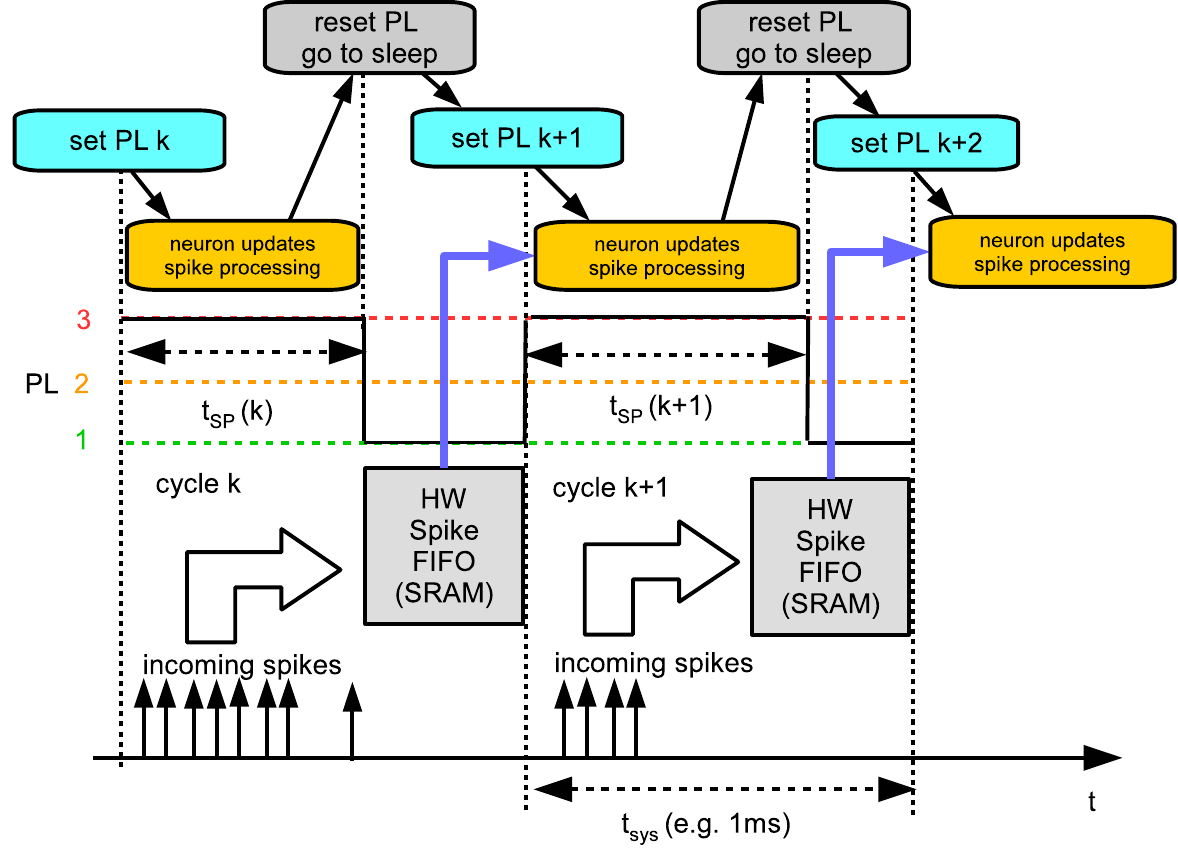}
  \caption{\textcolor{red}{Software flow for neuromorphic simulation with DVFS~\cite{Hoeppner2017}}}
	\label{fig:Neuro_DVFS_software_timeline}
\end{figure}

The computational load in neuromorphic simulations is determined by the neuron state updates and synaptic events. While the neuron processing cost is constant in each simulation cycle, the number of synaptic events to be processed per time and core strongly varies with network activity. Our approach for neuromorphic power management exploits this by periodically adapting the performance level to the current workload. Fig.~\ref{fig:Neuro_DVFS_software_timeline} visualizes the flow of a neuromorphic simulation with DVFS. Within a simulation cycle of length $t_\t{sys}$ spikes are received by the PE and registered in the hardware spike FIFO. While spikes of cycle $k$ are received those from cycle $k-1$ are processed without interrupting the processor at incoming spikes. At the beginning of each cycle $k$ the workload is estimated based on the spikes in the queue and the performance level is set to the lowest possible level that guarantees the completion of the neural processing within the cycle $t_\t{sys}$.

\subsection{Workload Estimation}
The workload for neuron processing is constant in each cycle and depends on the number of neurons per core $n_\t{neur}$ and the cost for each neuron state update $c_\t{neur}$ (in clock cycles):
\begin{align}
	c_\t{neur,tot} &= n_\t{neur}\cdot c_\t{neur}
\end{align}

Instead, the cost for synapse processing varies with the number of spike events
$l$ received and the fan-out $g_i$ of respective source neurons to the neurons
on this core (Fig.~\ref{fig:SNN_architecture_logical}).
The cost for decoding the synapse words and applying the weights to the input
buffers of the neurons is given by
\begin{align}
	c_\t{syn,tot} &= \sum_{l} g_{i(l)}\cdot c_\t{syn},
\end{align}
\textcolor{red}{where $c_\t{syn}$ is the cost for processing one synapse.}

In addition, there is a non-negligible overhead \textcolor{red}{$c_\t{pre-spike}$} for each received presynaptic
spike for looking up the synapse row address and the DMA transfer, which we
summarize as
\begin{align}
	c_\t{pre-spike,tot} &= l\cdot c_\t{pre-spike}
\end{align}

Hence, the total workload $c$ in clock cycles is given by
\begin{align}
	c&=c_\t{neur,tot} + c_\t{syn,tot} + c_\t{pre-spike,tot} + c_\t{other} \label{eq:total_workload}
\end{align}
where $c_\t{other}$ subsumes all remaining tasks such as the main experiment control or sending of spike events.

\subsection{Performance Level Selection}
\label{sec:plselection}
\textcolor{red}{
In each cycle $k$ the lowest perfomance level is chosen that achieves the
completion of neuron and synapse processing within time step $t_\t{sys}$.
The PL is determined by comparing the workload $c$
to thesholds $c_\t{th,1}$ and $c_\t{th,2}$
representing the compute capacities of PL1 and PL2 (in clock cyles per time step):
}
\begin{align}
	\t{PL}(k)=\begin{cases}
		\t{PL1},\t{if } c<c_\t{th,1}  \\
		\t{PL2}, \t{if } c_\t{th,1}\le c < c_\t{th,2}\\
		\t{PL3}, \t{if } c_\t{th,2}\le c
	\end{cases} \label{eq:PL_selection}
\end{align}


Then synaptic event processing and neuron state computation is performed at $\t{PL}(k)$.
When these tasks are completed after the spike processing time $t_\t{sp}(k)$
the processor is set back to PL1 and sleep mode (clock gating) is activated.
It reads
\begin{align}
	t_\t{sp}(k)&=c(k)\cdot f_{\t{PL},k}.
\end{align}
The optimization target for PL selection is to maximize $t_\t{sp}$ within a
single $t_\t{sys}$ period, since this relates to the usage of the minimum
required PL to complete the neuron and synapse processing tasks while
maintaining biological real-time operation.

To obtain a good estimate of the workload $c(k)$ at the beginning of the
simulation cycle, we must compute $c_\t{syn,tot}$ and iterate over all
spike events in the FIFO and add up their fan-outs which are implicitly
contained in the synapse row lookup table as the synapse row sizes.
This extra loop over the spike events creates an additional workload that
consumes part of the compute performance per simulation cycle and increases the
energy demands at the first glance.
Yet, the possibility to precisely adapt the performance level to the workload offers
great power saving capabilities, and it must be evaluated for each application
whether the detailed strategy (Eq.~\ref{eq:PL_selection}) pays off.

\begin{figure}[htb]
	\centering
		\includegraphics{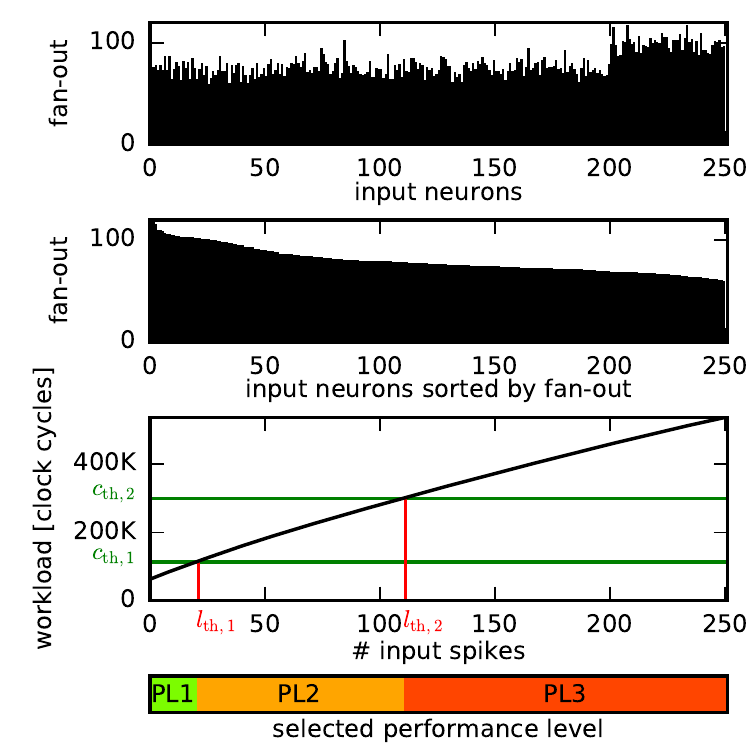}
  \caption{Secure strategy for selection of \textcolor{red}{performance} level thresholds:
	Depending on the fan-out of the input neurons per core, the worst-case
	workload for $l$ received spikes can be calculated assuming that the spikes
	belong to the $l$ inputs with the largest fan-out.
	The PL thresholds $l_\t{th,1}$, $l_\t{th,2}$ are obtained by the
	intersections with the compute capacities $c_\t{th,1}$, $c_\t{th,2}$ of the
	PLs.
  \textcolor{red}{The PL is selected according to Eq.~\ref{eq:PL_selection_simple}.}
	The strategy is illustrated for one core of the synfire chain model
	investigated in Sec.~\ref{sec:synfire_chain}.
	}
	\label{fig:pl_selection_worstcase}
\end{figure}

For this paper, however, we consider a simple performance level
selection model based on the number of received spikes $l$ in the spike queue:

\textcolor{red}{For this paper, however, we consider a simple performance level
selection model that compares the number of received spikes $l$ in the spike
queue with thresholds $l_\t{th,1}$ and $l_\t{th,2}$:
}
\begin{align}
	\t{PL}(k)=\begin{cases}
		\t{PL1},\t{if } l<l_\t{th,1}  \\
		\t{PL2}, \t{if } l_\t{th,1}\le l < l_\t{th,2}\\
		\t{PL3}, \t{if } l_\t{th,2}\le l
	\end{cases} \label{eq:PL_selection_simple}
\end{align}
\textcolor{red}{The relation between selected performance level and the thresholds can be seen at the bottom of Fig.~\ref{fig:pl_selection_worstcase}.}
This strategy was used in the preceding work~\cite{Hoeppner2017} and allows a fast selection of the performance level as $l$ is immediately available at the
beginning of the cycle $k$. In turn, the thresholds $l_\t{th,1}$ and $l_\t{th,2}$ must be tuned for each
application for the best exploitation of energy savings.
Care has to be taken that all spike events can be processed at the chosen PL
before the end of the time step.
To ensure this, we employ a worst-case strategy for setting the performance level
thresholds based on the fan-out of source neurons per core, as illustrated in
Fig.~\ref{fig:pl_selection_worstcase}:
For $l$ spikes received, the worst case for the workload is when these spikes
belong to the $l$ input neurons with the highest fan-out.
As we know the fan-out of each neuron before the simulation, we can set the
perfomance level thresholds according to this worst case:
For this, we sort the fan-out of all inputs per core in descending order, and
compute the worst-case workload for $l$ spikes received according to
Eq.~\ref{eq:total_workload}.
Then, the thresholds $l_\t{th,1}$ and $l_\t{th,2}$ are given by the
intersection points of the worst-case workload and the compute capacities
$c_\t{th,1}$ and $c_\t{th,2}$, as sketched in the bottom of
Fig.~\ref{fig:pl_selection_worstcase}.
The advantage of this approach is two-fold: On the one hand, it guarantees the
completion of synapse processing within the time step, on the other hand it is
universal and can be employed to any network architecture.
In real applications, however, higher thresholds might suffice when the
worst-case that the $l$ input neurons with the highest fan-out fire at the same
time never occurs.

%% file: pm_theory.tex
%
%
To derive a model for power consumption and energy efficiency~\cite{Haid2010} of the PE within the many core system, a breakdown into the individual contributors (from application perspective) is required. Based on the definitions from~\cite{Stromatias2013}, the total PE power consumption is split into baseline power, \textcolor{red}{and the energies for neuron and synapse processing in the simulation cycles, as explained in the following. The parameter extraction is shown in Sec.~\ref{sec:pmextraction}}.

\subsection{Baseline Power}
The baseline power is consumed by the processor running the neuromorphic simulation kernel without processing any neurons or synapses. The $t_\t{sys}$ timer events are received but trigger no neuron processing or synapse processing of the PEs. The baseline power also includes the PE leakage power, when connected to the particular $V_\t{DD}$ rails of the PL. The baseline power of PL $i$ is $P_{\t{BL},i}$.

\subsection{Neuron Processing \textcolor{red}{Energy}}
\textcolor{red}{The neuron processing energy per simulation time step at PL $i$ is modeled by,
\begin{align}
	E_{\t{neur},i}=E_{\t{neur},0,i}+ e_\t{neur,i}\cdot n_\t{neur}
\end{align}
assuming a linear relation with an offset energy $E_{\t{neur},0,i}$ (as extracted in Sec.~\ref{sec:pmextraction}) and an incremental neuron processing energy $e_\t{neur,i}$ per neuron. The total energy depends on the number of neurons $n_\t{neur}$ mapped to the particular PE, independent from the network activity.}

\subsection{Synapse Processing \textcolor{red}{Energy}}
\textcolor{red}{The synapse energy is the PE contribution caused by processing synaptic events within each simulation time step. The PE synapse energy at PL $i$ is modeled by,
\begin{align}
	E_{\t{syn},i}=E_{\t{syn},0,i}+ e_\t{syn,i}\cdot n_\t{syn}
\end{align}
assuming a linear relation with an offset offset energy $E_{\t{syn},0,i}$ and an incremental energy $e_\t{syn,i}$ per synaptic event. The total energy depends on the number of synaptic events  $n_\t{syn}$ within a simulation cycle on the particular PE, and thereby from the network activity.}

\bigskip
From these definitions, the total energy consumed within a simulation cycle $k$ of length $t_\t{sys}$ at PL $i$ reads
\textcolor{red}{
\begin{align}
	E_{\t{cycle}}(k)=&P_{\t{BL},i}\cdot t_\t{sp} +P_{\t{BL},1}\cdot(t_\t{sys}-t_\t{sp}) \nonumber \\
	&+E_{\t{neur},0,i}+e_\t{neur,i}\cdot n_\t{neur}  \nonumber \\
	&+E_{\t{syn},0,i}+e_\t{syn,i}\cdot n_\t{syn} \label{eq:esumcycle}
\end{align}
}
assuming that after the completion of neuron and synapse processing within that cycle after $t_\t{sp}$ the PE is set back to PL1 (as illustrated in Fig.~\ref{fig:Neuro_DVFS_software_timeline}). This reduces baseline power in that idle time of length $t_\t{sys}-t_\t{sp}$. The average power consumption over the total experiment of $k_\t{max}$ cycles reads
\begin{align}
	P_\t{avg}&=\frac{1}{k_\t{max}\cdot t_\t{sys}}\cdot\sum\limits_{k=1}^{k=k_\t{max}} E_{\t{cycle}}(k)
\end{align}

%% file: testchip_santos.tex
%
%

\subsection{Overview}

\begin{figure}[htb]
	\centering
		\includegraphics[width=0.47\textwidth]{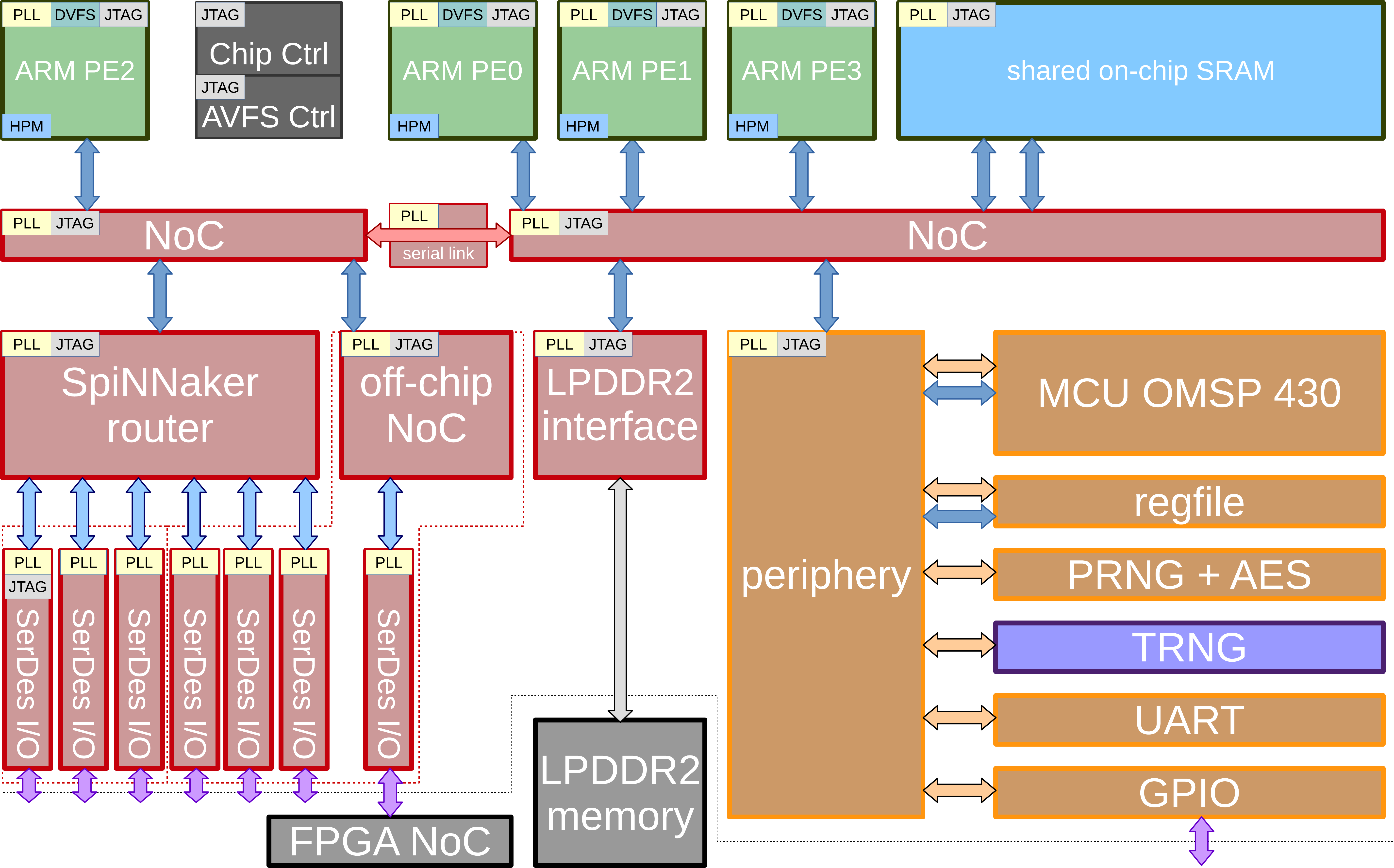}
	\caption{Test chip block diagram}
	\label{fig:Santos28_Blockdiagram}
\end{figure}

\begin{figure}[htb]
	\centering
		\includegraphics[width=0.38\textwidth]{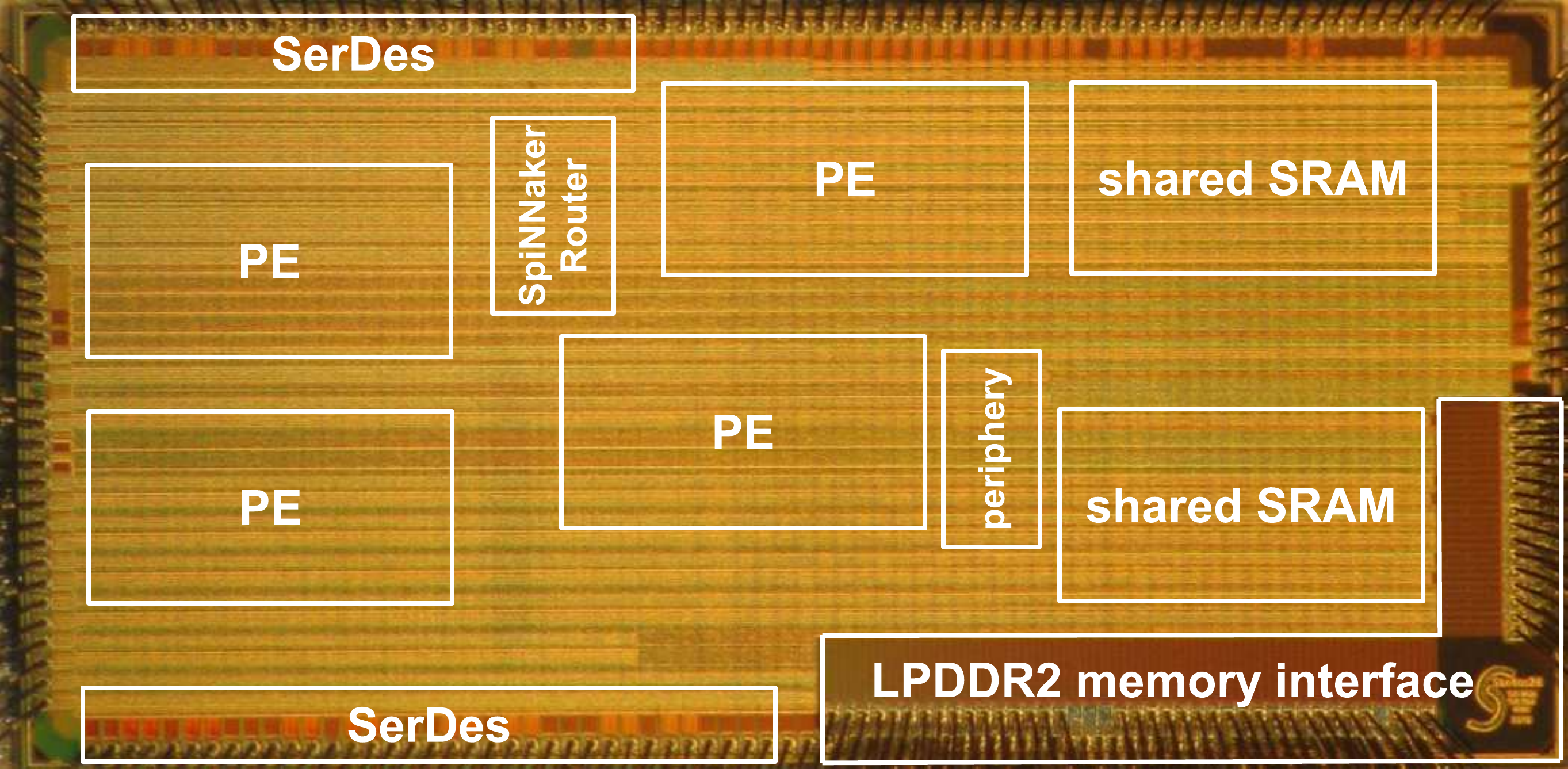}
	\caption{Chip Photo, with major building blocks marked~\cite{Hoeppner2017}}
	\label{fig:Santos_chip_photo}
\end{figure}

\begin{figure}[htb]
	\centering
		\includegraphics[width=0.38\textwidth]{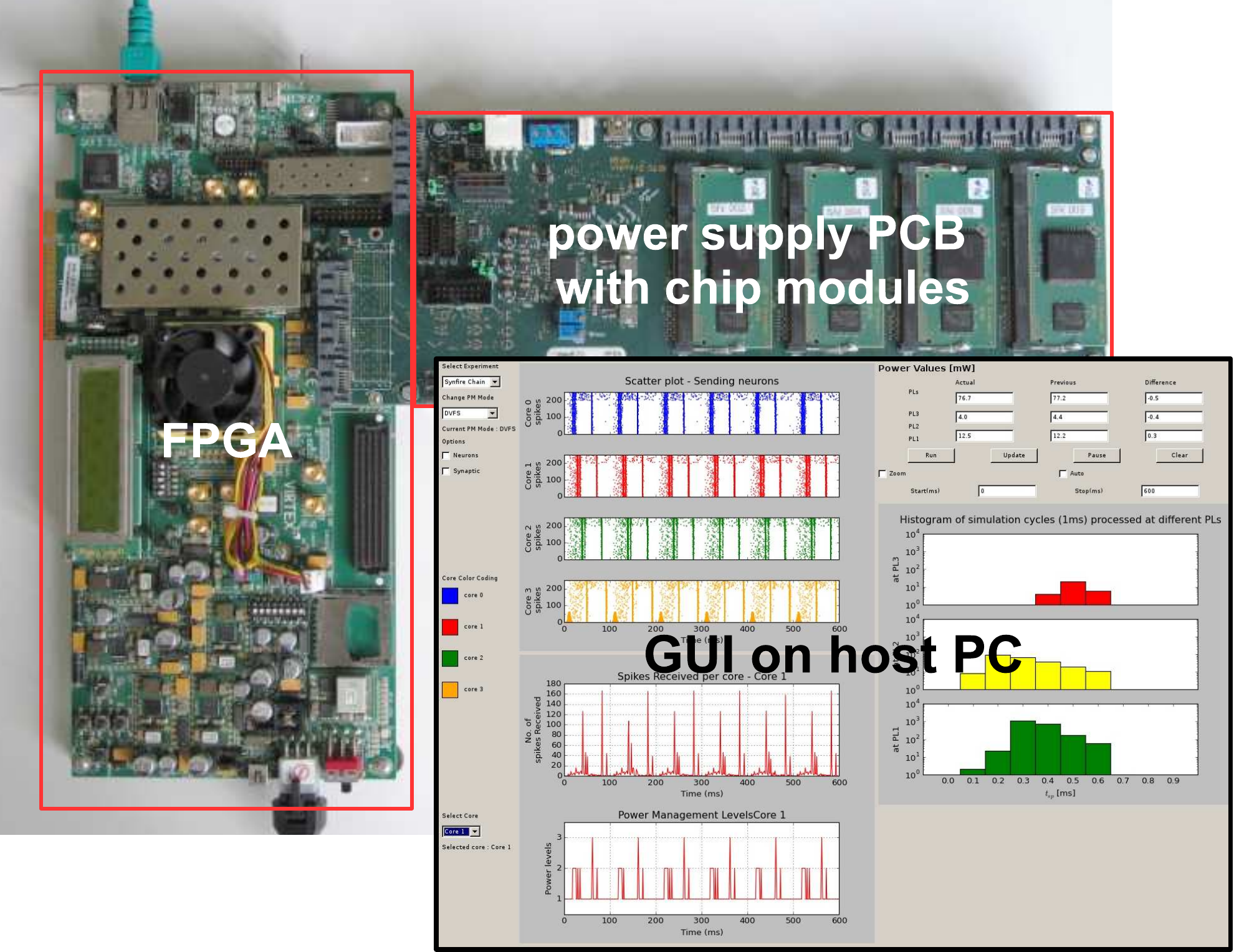}
  \caption{Measurement setup, PCB with power supply and up to 4 test chip modules, FPGA board and host PC GUI~\cite{hoeppner2017live}}
	\label{fig:Santos_setup_photo}
\end{figure}

A test chip of the SpiNNaker2 neuromorphic many core system has been implemented in GLOBALFOUNDRIES \SI{28}{nm} SLP CMOS technology. Its block diagram is shown in Fig.~\ref{fig:Santos28_Blockdiagram}. It contains 4 PEs with ARM M4F processors and hardware accelerators for exponentials~\cite{Partzsch2017b} and true random number generators~\cite{Neumarker2016}. Each PE includes \SI{128}{kB} local SRAM and the proposed power management architecture with three PLs. \SI{128}{MByte} off-chip DRAM is interfaced by LPDDR2. All on-chip components are connected by a NoC, where the longer range point to point connections are realized using the serial on-chip link~\cite{Hoeppner2015}. The chip photo is shown in Fig.~\ref{fig:Santos_chip_photo}. For lab evaluation a power supply PCB is used which hosts up to 4 chip modules. This is connected to an FPGA evaluation board via SerDes links which then connects to the host PC via standard Ethernet. The setup is shown in Fig.~\ref{fig:Santos_setup_photo}, consisting of a power supply PCB hosting up to 4 chip modules and an FPGA board for host PC communication over Ethernet. A graphical user interface running on the host PC allows exploration of the DVFS measurements \cite{hoeppner2017live}.

\subsection{PE Measurement Results}
\label{sec:pemeasresults}

\begin{figure}[htb]
\centering
\subfigure[maximum PE clock frequency \label{fig:santos_arm_shmoo_vdd_freq}]{\includegraphics[width=0.40\textwidth]{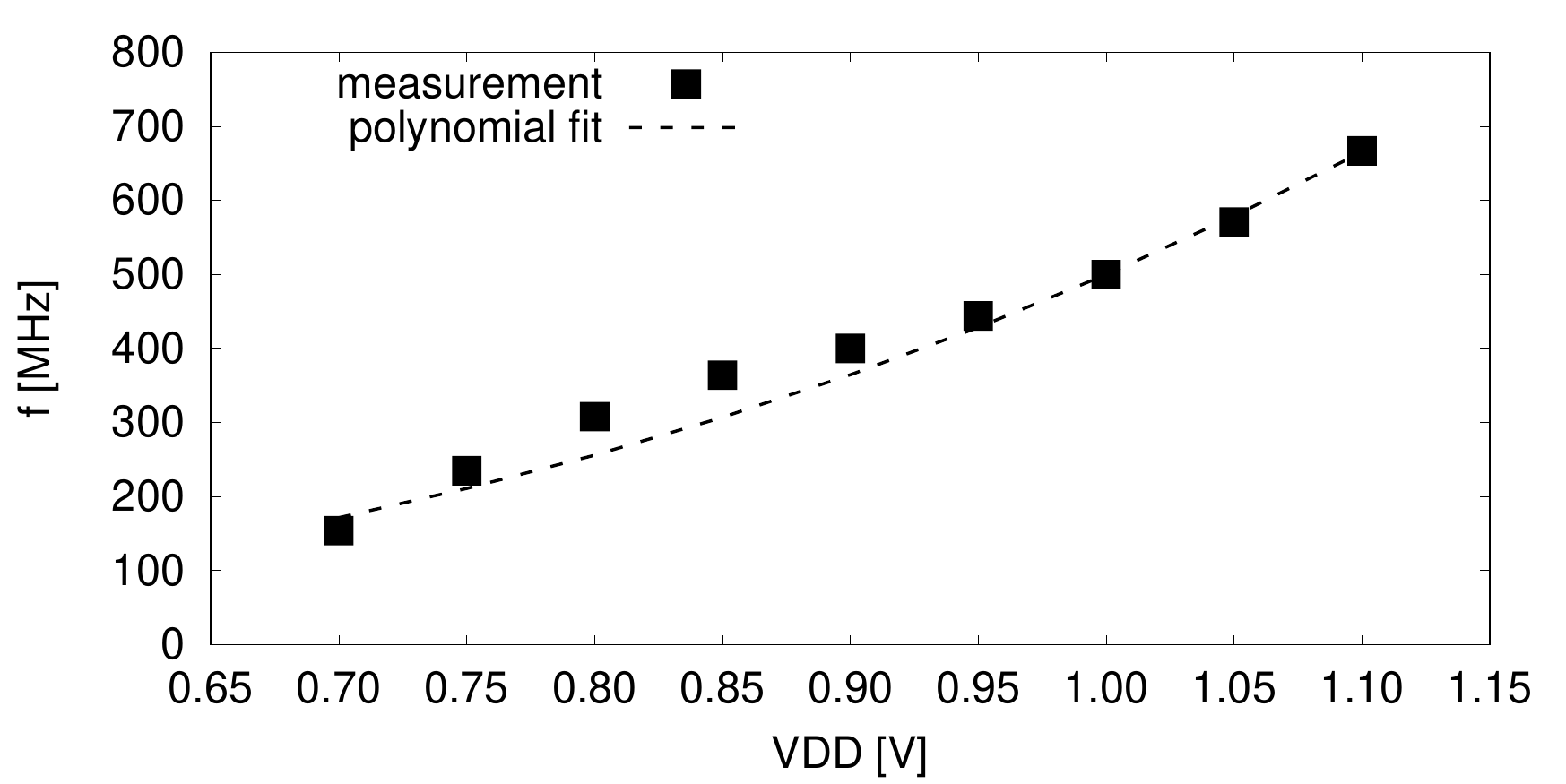}}
\subfigure[power and energy per operation \label{fig:santos_arm_dvfs_power_energy}]{	\includegraphics[width=0.40\textwidth]{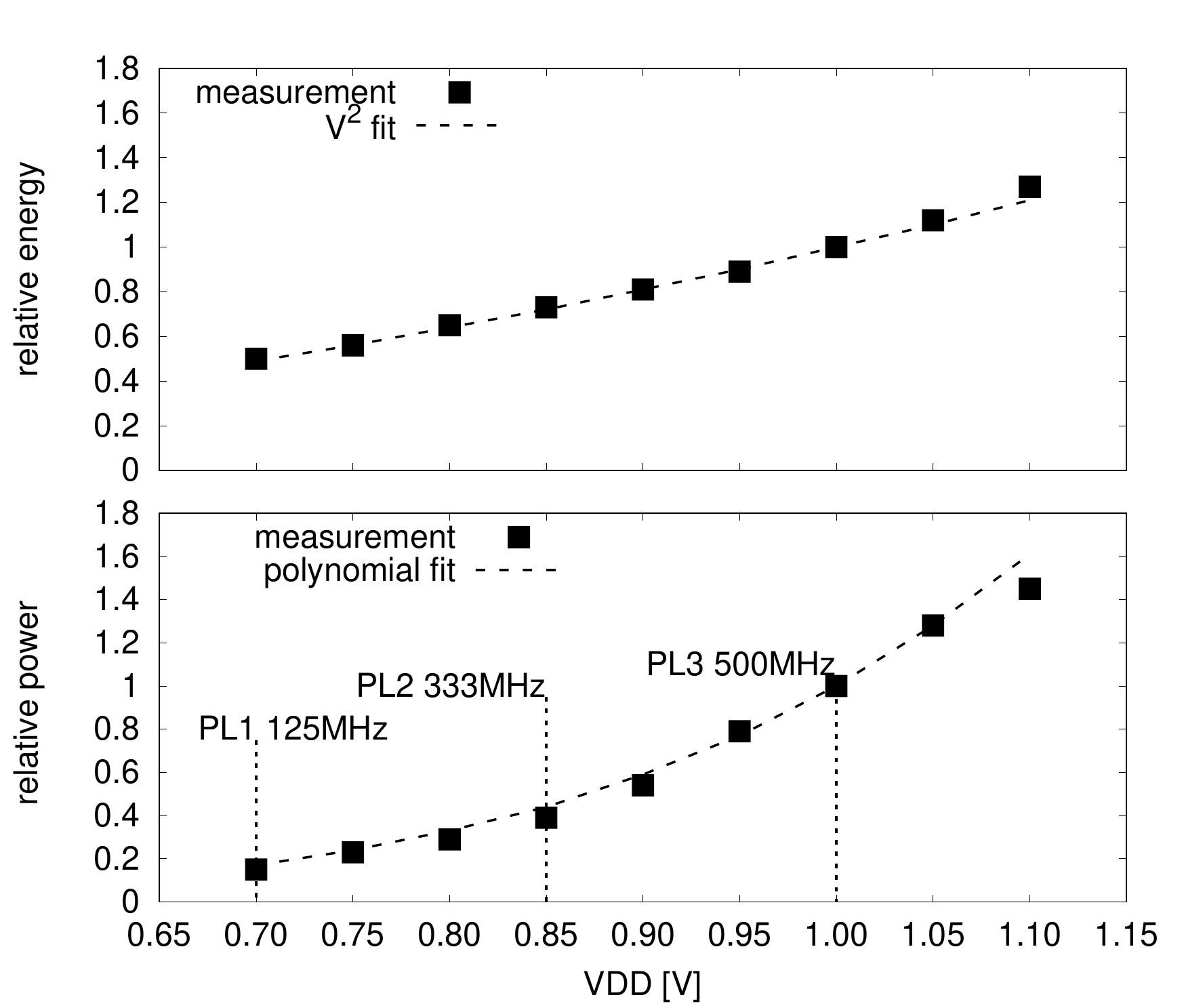}}
\caption{Processing element (ARM M4F) DVFS measurements}
	\label{fig:santos_dvfs_meas}
\end{figure}

Fig.~\ref{fig:santos_arm_shmoo_vdd_freq} shows the measured maximum PE frequency versus the supply voltage $V_\t{DD}$ including a polynomial fitting curve. The PE is operational with high yield (including SRAM) down to \SI{0.7}{V}. \textcolor{red}{In this testchip the three PLs for the neuromorphic application are defined as PL1 (\SI{0.70}{V}, \SI{125}{MHz}), PL2 (\SI{0.85}{V}, \SI{333}{MHz}) and PL3 (\SI{1.00}{V}, \SI{500}{MHz}) spanning a range from the nominal supply voltage of this process node of \SI{1.00}{V} down to the minimum SRAM supply of \SI{0.70}{V}. The design has been implemented for target performance at PL3. Hold timing has been fixed in all corners. The achieved performance at PL1 and PL2 has been analyzed from sign-off timing analyses.}  Fig.~\ref{fig:santos_arm_dvfs_power_energy} shows the scaling of the PE energy-per-task metric and power consumption when scaling $V_\t{DD}$ and $f$. The dynamic energy consumption of the operating processor, consisting of internal and switching power of logic gates and interconnect structures dominates in this design implementation. Therefore a fitting of $E_\t{task}=e_\t{task,norm} \cdot V_\t{DD}^2$, where $e_\t{task,norm}$ is the normalized energy per specific task, is applicable here. By voltage and frequency scaling in the mentioned ranges, the energy consumption per task can be reduced by \SI{50}{\%} and the power consumption by \SI{85}{\%} relative to the nominal operation point of PL3 (\SI{1.00}{V},\SI{500}{MHz}).

\begin{figure*}[htb]
\centering
\subfigure[PU to \SI{0.7}{V}  \label{fig:shmoo_pu_0_70_125}]{ 	\includegraphics[width=0.39\textwidth]{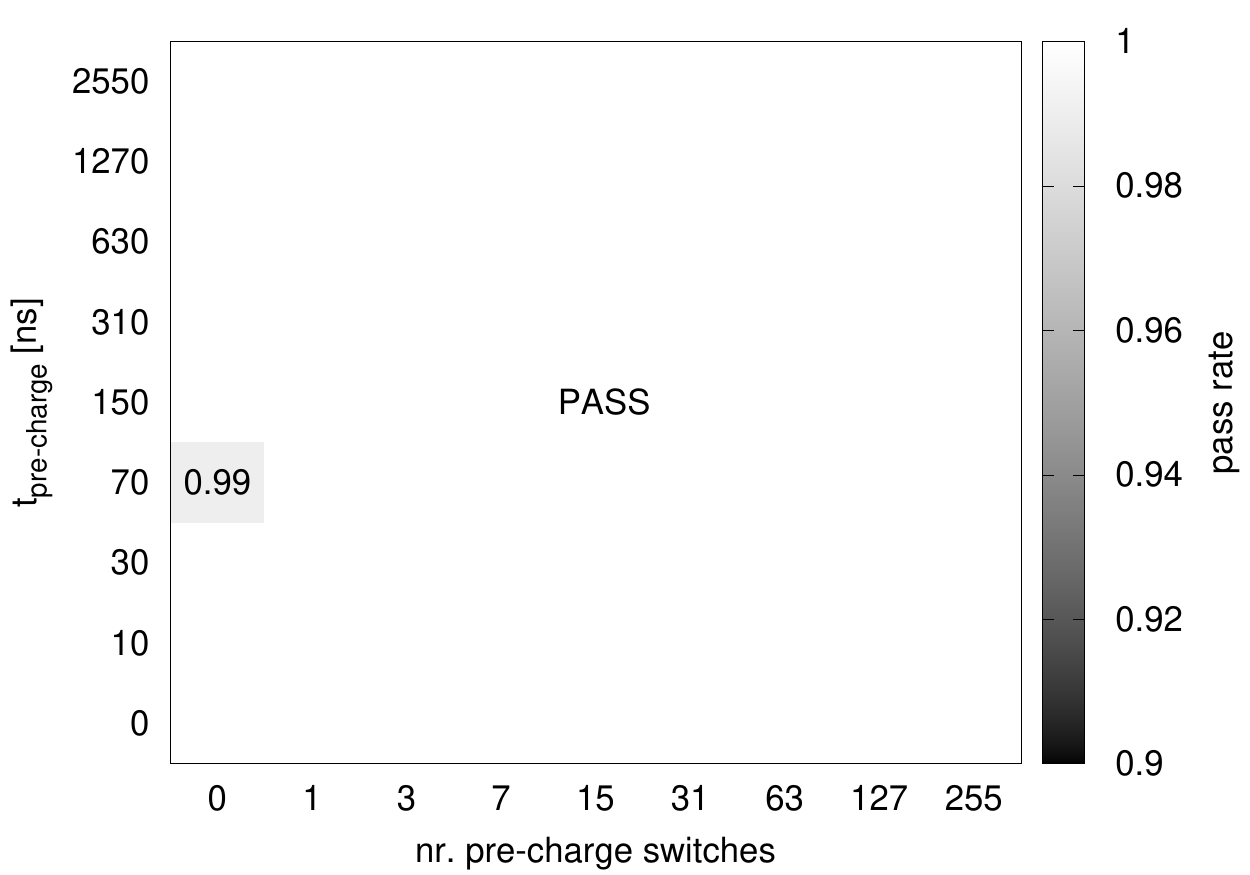}}
\subfigure[PU to \SI{0.85}{V}  \label{fig:shmoo_pu_0_85}]{ 	\includegraphics[width=0.39\textwidth]{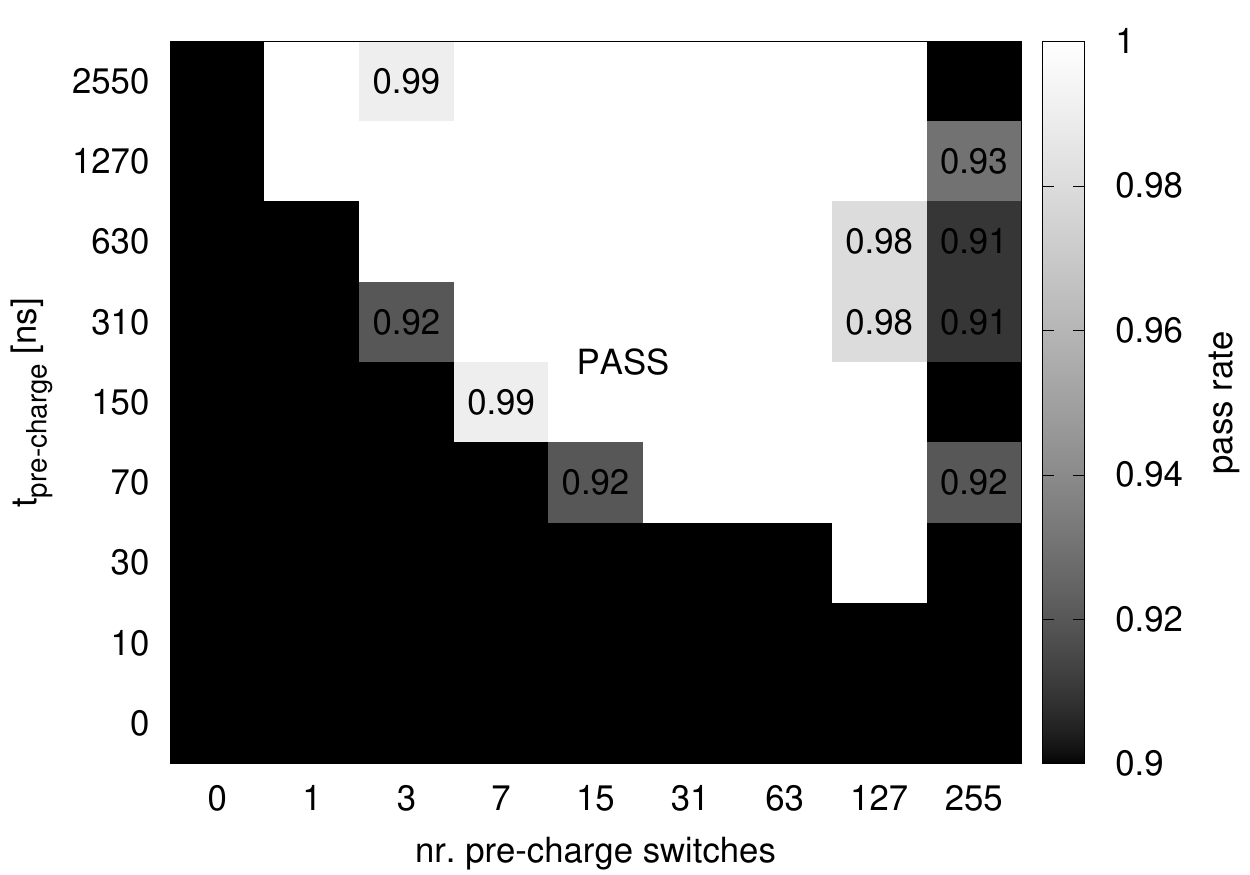}}
\subfigure[PU to \SI{1.0}{V} (always-on-rail)  \label{fig:shmoo_pu_1_00}]{ 	\includegraphics[width=0.39\textwidth]{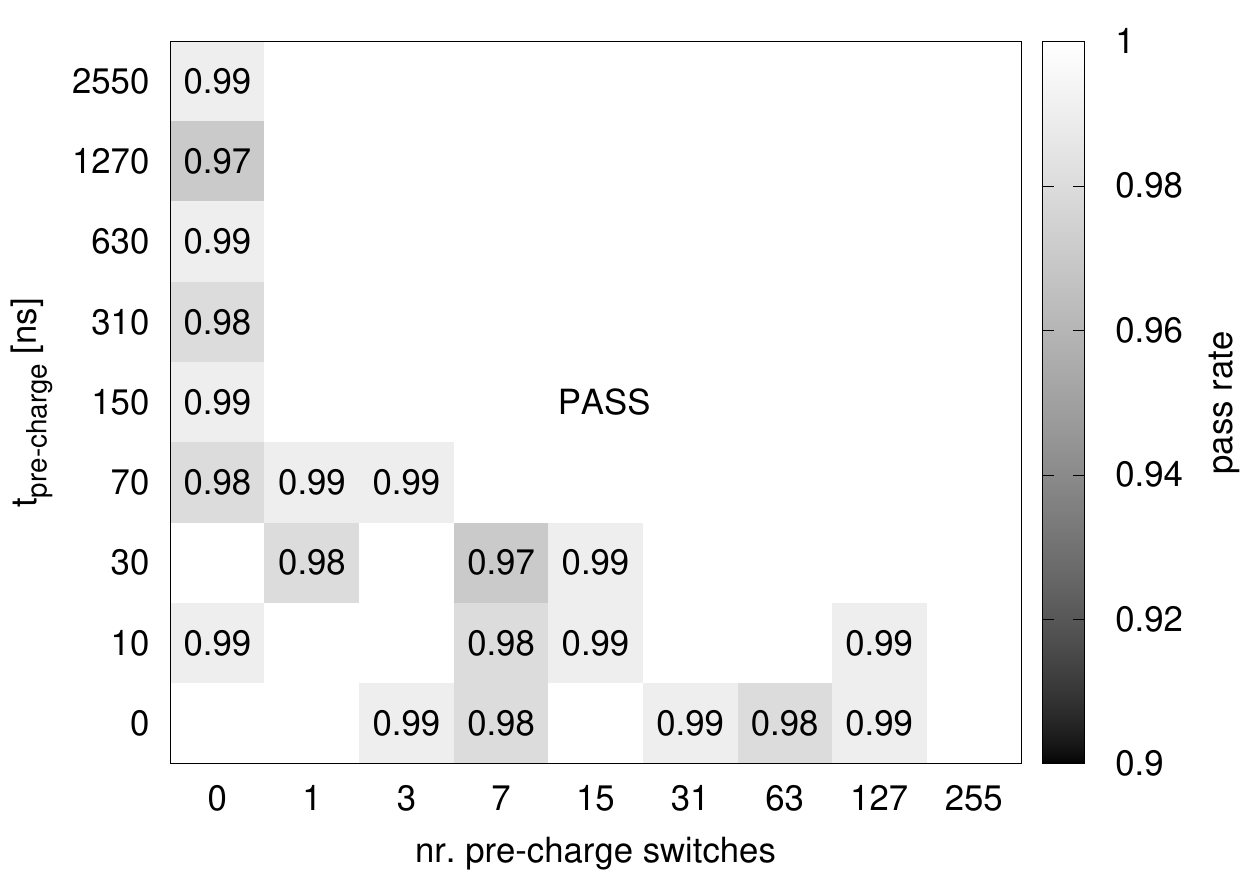}}
\subfigure[SC from \SI{0.7}{V} to \SI{0.85}{V} \label{fig:shmoo_sc_1}]{ \includegraphics[width=0.39\textwidth]{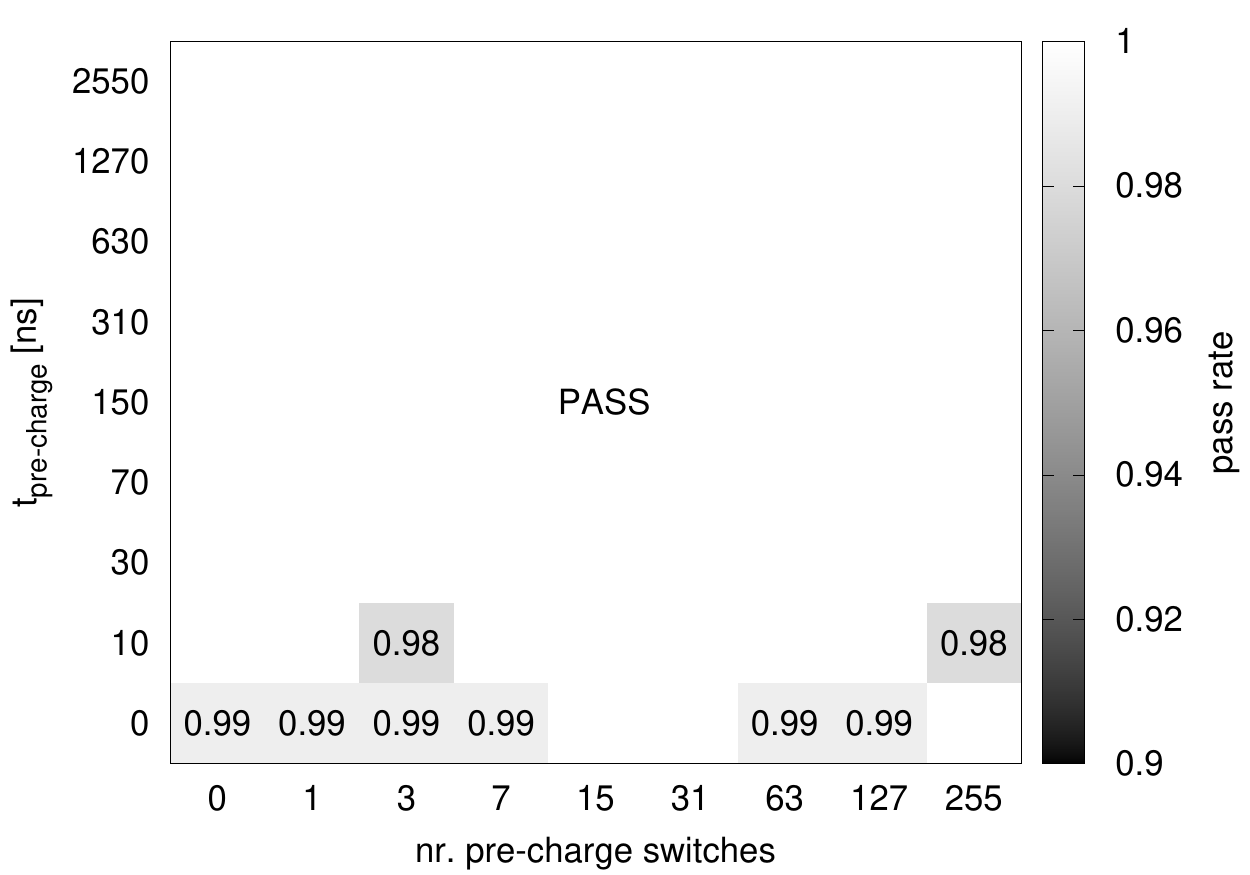}}
\caption{Robustness measurements of PL changes, color maps show the pass-rate of a software test case running of the victim PE, when the aggressor PE is switching its PL, parameters are the number of pre-charge switches and the pre-charge time.}
	\label{fig:pl_switching_shmoo_plots}
\end{figure*}

The robustness of the PL switching has been analyzed by measurements in which a particular PE (aggressor) performs PL switches to a target supply net while another PE (victim) is actively operating at this target net, performing a task which can be checked for successful execution. The PL switching is repeated multiple times. The experiment is repeated for various settings of supply net pre-charge time and number of activated pre-charge switches (see Sec.~\ref{sec:powermanagementhardwarearchiecture}). Fig.~\ref{fig:pl_switching_shmoo_plots} shows the measurement results for various scenarios for power-up (PU) and supply (PL) change (SC).

In Fig.~\ref{fig:shmoo_pu_0_70_125} PU to the \SI{0.7}{V} rail (PL1) is measured, which is very robust versus the switching time and the number of pre-charge switches. This is caused by the fact that the \SI{125}{MHz} frequency setting of PL1 has significant headroom to the maximum possible frequency of \SI{166}{MHz} (see Fig.~\ref{fig:santos_arm_shmoo_vdd_freq}), thereby tolerating temporary supply drops during switching.

In case of PU to PL2 (Fig.~\ref{fig:shmoo_pu_0_85}), a strong dependency of the robustness on the pre-charge time and number of switches is visible. For a large number of pre-charge switches the victim PE fails due to the large rush current induced voltage drop at PL2. A smaller number of pre-charge switches anyway requires a minimum pre-charge time, which decreases with increasing number of switches. This is caused by the fact that if the pre-charge time is too short, all power switches are activated although the PE supply net has not yet settled to PL2, resulting in a significant rush current and PL2 supply drop directly after the pre-charge phase. The PU behavior to PL3 is similar (Fig.~\ref{fig:shmoo_pu_1_00}) but much more robust, since the \SI{1.0}{V} supply net PL3 is the always on-domain of the chip toplevel, therefore having high on-chip decoupling capacitance which tolerates larger rush currents during power switching. Fig.~\ref{fig:shmoo_sc_1} shows the SC scenario from PL1 to PL2 which is most critical (PL2 has smaller on-die decoupling capacitance than PL3). Also high switching robustness is achieved here. In summary, for safe operation well within the PASS regions of Fig.~\ref{fig:pl_switching_shmoo_plots}, a setting of 31 pre-charge switches enables save PU in $\approx 1\mu\t{s}$ and SC in $< \SI{100}{ns}$.

\subsection{DVFS Power Model Parameter Extraction}
\label{sec:pmextraction}

\begin{figure}[htb]
	\centering
		\includegraphics[width=0.40\textwidth]{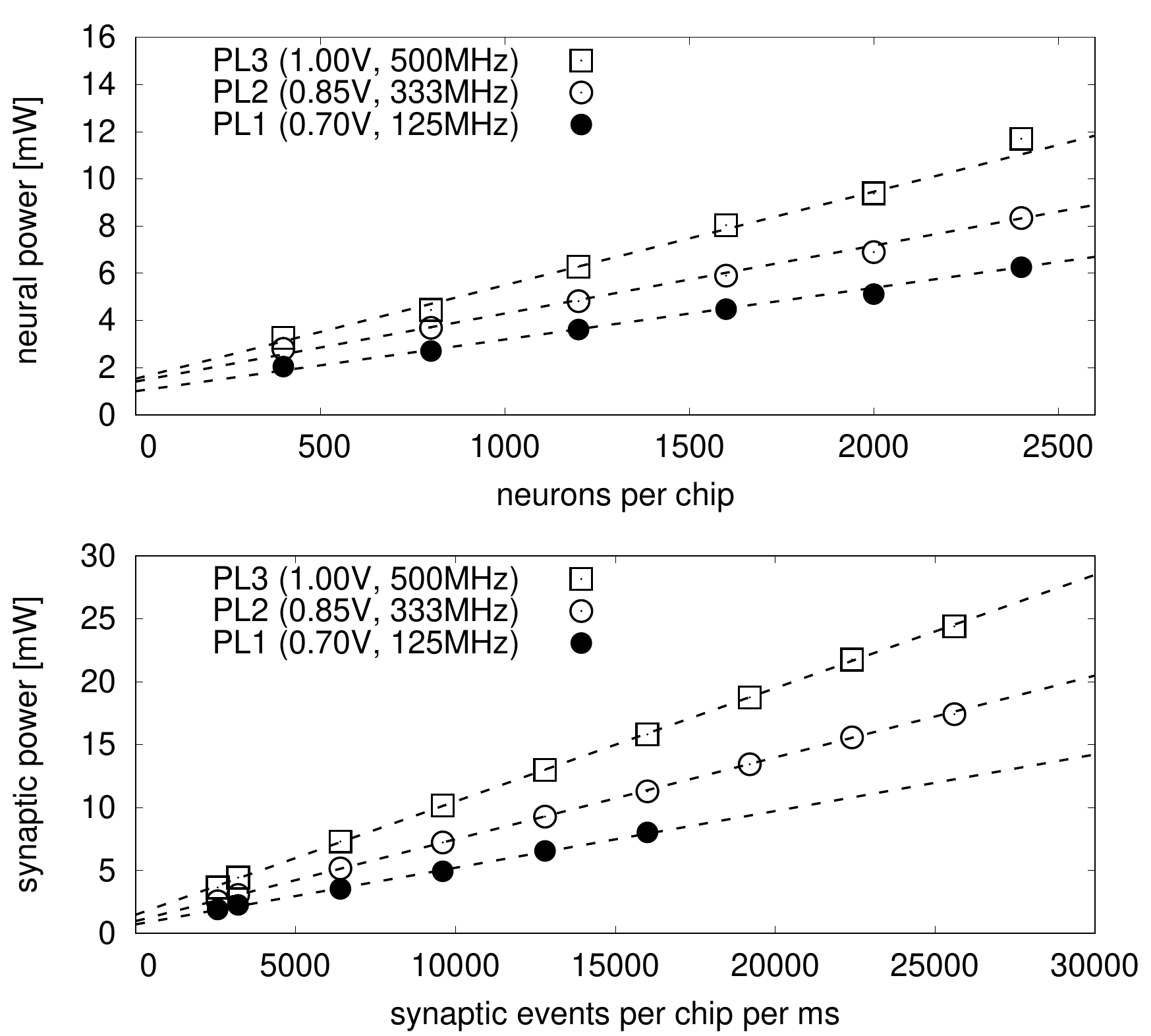}
  \caption{Neural and synapse energy parameter extraction. The data points at 16,000 synaptic events per ms in the bottom plot also show up in Table~\ref{tab:powermeas} as power figures for the locally-connected network.}
	\label{fig:santos_arm_dvfs_neuron_synapse_energy}
\end{figure}

The power management model parameters from Sec.~\ref{sec:pmtheory} have been extracted. For baseline power $P_{\t{BL}}$ extraction the simulation kernel without neuron processing code and synapse processing code has been executed at different PLs, including the $t_\t{sys}$ trigger to the PE. The neuron processing power has been determined by running the neuron state calculation for different numbers of leaky integrate-and-fire (LIF) neurons with conductance-based synapses as shown in Fig.~\ref{fig:santos_arm_dvfs_neuron_synapse_energy}. The power for synapse processing has been measured by running the synapse calculation code with a varying number of synaptic events.
Therefore, we used a locally-connected network with 80 neurons per core that are connected to all neurons on the same core, but not to other cores. Then we modified the number of spiking neurons per core and time step to vary the number of synaptic events. This network topology is identical to the local network used in~\cite{Stromatias2013} and allows the direct comparison of energy efficiency to the first generation SpiNNaker system in Sec.~\ref{sec:resultscomparison}.

The results of the power model parameter extraction are shown in Fig.~\ref{fig:santos_arm_dvfs_neuron_synapse_energy}. Linear interpolation has been applied to extract the model parameters for neuron \textcolor{red}{energy $E_{\t{neur},0}=P_{\t{neur},0}\cdot t_\t{sys}$} and $e_\t{neur}$ and synapse \textcolor{red}{energy $E_{\t{syn},0}=P_{\t{syn},0}\cdot t_\t{sys}$} and $e_\t{syn}$, respectively. All parameters are summarized in Tab.~\ref{tab:npmparameters}.
As expected, the energy per task ($e_\t{neur}$ or $e_\t{syn}$) scales with $V_\t{DD}^2$, while the baseline power at PL1 is only \SI{20}{\%} of the power at PL3, which is consistent with the results in Sec.~\ref{sec:pemeasresults}.

\setrowcolors{}

\begin{table}[htb]
  \begin{minipage}{0.47\textwidth}
    \centering
    \renewcommand{\arraystretch}{1.1}
    \caption{Measured parameters of power management model}
    \label{tab:npmparameters}
    \centering
    \footnotesize
    \begin{tabular}{lccc} \toprule
                                 & PL1 (\SI{0.70}{V}) & PL2 (\SI{0.85}{V}) & PL3 (\SI{1.00}{V}) \\ \midrule
      $P_{\t{BL,leak}}$ [\si{mW}]& 8.94               & 20.03              & 28.53          \\
      $P_{\t{BL}}$ [\si{mW}]     & 14.92              & 37.44              & 71.17              \\
      \textcolor{red}{$E_{\t{neur},0}$ [\si{nJ}]} & 1000               & 1410               & 1540               \\ 
      $e_\t{neur}$    [\si{nJ}]  & 2.19               & 2.88               & 3.96               \\
      \textcolor{red}{$E_{\t{syn},0}$ [\si{nJ}]}  & 730               & 990               & 1490               \\ 
      $e_\t{syn}$   [\si{nJ}]    & 0.45               & 0.65               & 0.90               \\
      \bottomrule

    \end{tabular}
  \end{minipage}
\end{table}

%% file: results.tex
\subsection{Benchmark Networks}
To show the capabilities of neuromorphic power management we implement three diverse spiking neural networks for execution on the chip.
The benchmarks were selected to cover the wide range of neural activity found in the brain~\cite{Buzsaki2014}, ranging from asynchronous irregular activity over synchronous spike packets to network bursts.
In all networks we use leaky integrate-and-fire neurons with conductance-based synapses.
Table~\ref{tab:networks} lists characteristics of the networks and the
thresholds $l_\t{th}$ of received spikes used for the performance level selection.
The thresholds were determined using the worst-case approach described in
Section~\ref{sec:plselection}, except for the synfire network, which uses the
same numbers as in the preceding work~\cite{Hoeppner2017} for consistency.

\begin{figure}[htb]
	\centering
		\subfigure[Synfire chain network~\cite{Hoeppner2017}\label{fig:synfire_chain}]{\includegraphics[width=0.47\textwidth]{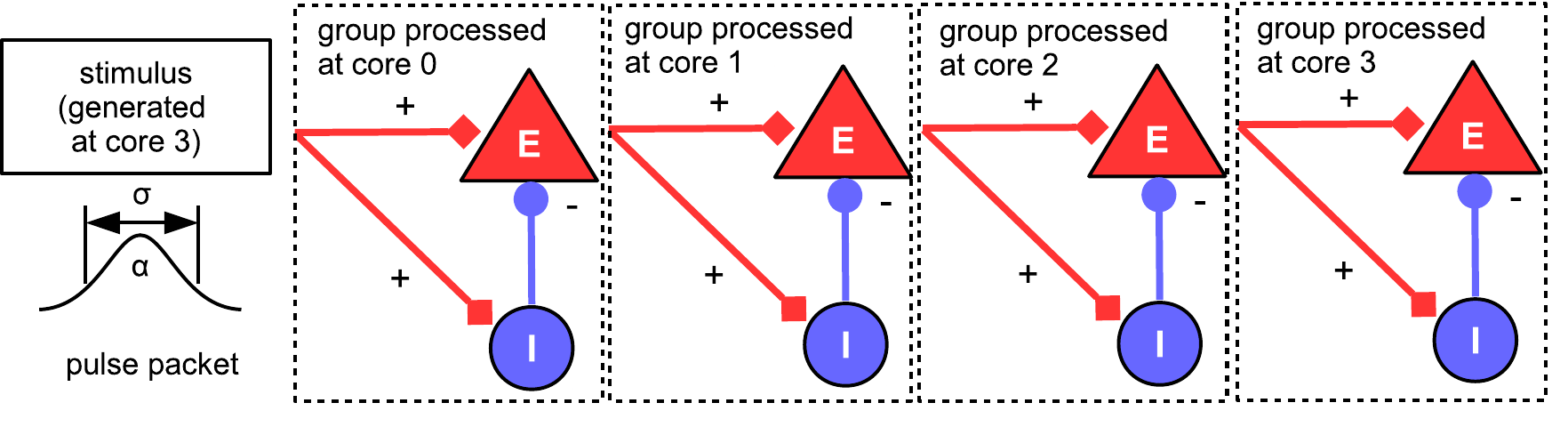}}
		\subfigure[Bursting network\label{fig:bursting_network}]{\includegraphics[width=0.2\textwidth]{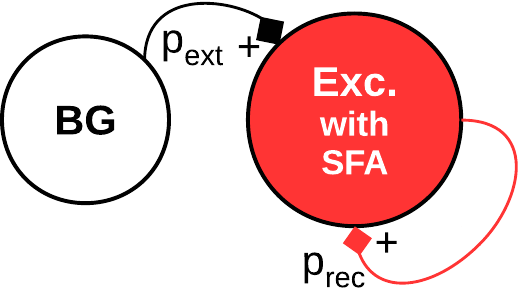}}
		\hspace{5mm}
		\subfigure[Asynchronous irregular firing network\label{fig:ai_network}]{\includegraphics[width=0.2\textwidth]{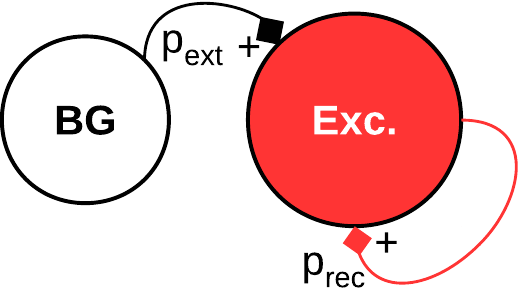}}
	\caption{Benchmark networks}\label{fig:benchmark_network}
\end{figure}

\setrowcolors{}
\begin{table}[htb]
  \begin{minipage}{0.47\textwidth}
    \centering
    \renewcommand{\arraystretch}{1.1}
    \caption{Benchmark networks}\label{tab:networks}
    \centering
    \footnotesize
    \begin{tabular}{lccc} \toprule
                        & synfire     & bursting & async \\ \midrule
      neuron model      & LIF + noise & LIF      & LIF   \\
      neurons per core  & 250         & 250      & 250   \\
      synapses per core & 20000       & 25000    & 10000 \\  
      avg.\ fan-out     & 80          & 21       & 8     \\
      $l_\t{th,1}$      & 20          & 47       & 47    \\
      $l_\t{th,2}$      & 100         & 214      & 229   \\
      \bottomrule{}
    \end{tabular}
  \end{minipage}
\end{table}

\subsubsection{Synfire Chain}
\label{sec:synfire_chain}


\begin{figure}[htb]
	\centering
		\includegraphics[width=0.47\textwidth]{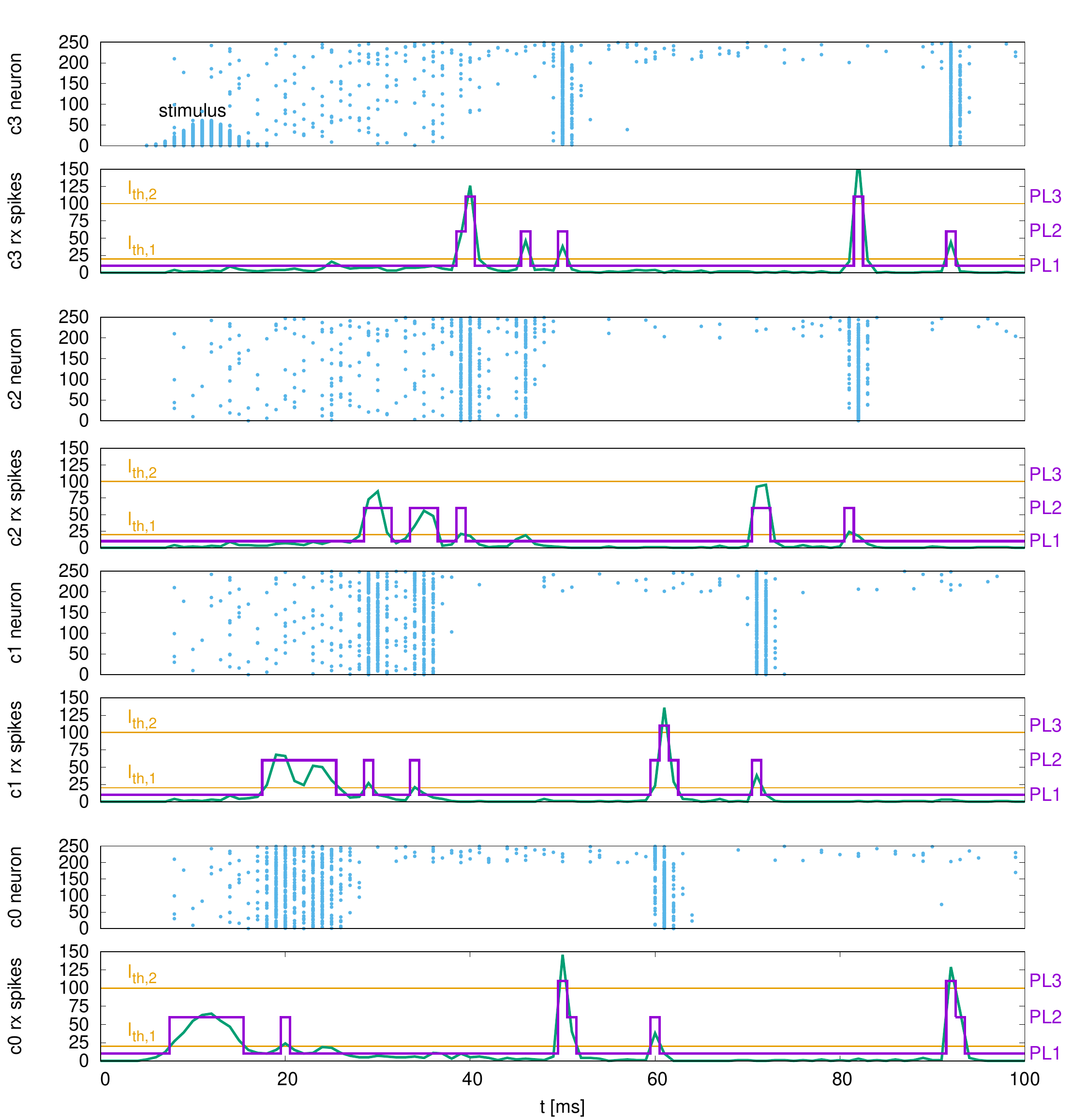}
  \caption{Synfire chain benchmark spike train, sent spikes (blue), number of received spikes per core (green), PL selection thresholds (orange) and core PL (violet), plot shows zoom in to \SI{100}{ms} simulation time for illustration of fast PL changes.}\label{fig:synfire_chain_spikes_pl}
\end{figure}

The first benchmark is a synfire chain network~\cite{Kremkow2010} which was already used in the preceding work~\cite{Hoeppner2017}: Synfire chains are feedforward networks that propagate synchronous firing activity through a chain of neuron groups~\cite{Abeles2009}. We implement a synfire chain with feedforward inhibition~\cite{Kremkow2010} consisting of \SI{4}{groups} (Fig.~\ref{fig:synfire_chain}), each with \SI{200}{excitatory} (E) and \SI{50}{inhibitory} (I) neurons.
As in~\cite{Kremkow2010} the neurons receive a normally distributed noise current.
Excitatory neurons are connected to both excitatory and inhibitory neurons of the next group, while inhibitory neurons only connect to the excitatory population of the same group.
There are \SI{25}{presynaptic} connections per neuron from I to E and \SI{60}{presynaptic} connections per neuron from E of the previous group, the delays are \SI{8}{ms} within a group and \SI{10}{ms} between groups.
There are no recurrent connection within a population. We simulate one group per core and connect the last group to the first one. At start, the first group receives a Gaussian stimulus pulse packet generated on core 3 (\SI{400}{spikes}, $\sigma=\SI{2.4}{ms}$). As shown in Fig.~\ref{fig:synfire_chain_spikes_pl}, the pulse packet propagates stably from one group to another, where the feedforward inhibition ensures that the network activity does not explode.


As shown in Fig.~\ref{fig:synfire_chain_spikes_pl}, cores adapt their PLs to the number of incoming spikes within the current \SI{1}{ms} simulation cycle. Fig.~\ref{fig:pl_time_done_hist} shows histograms of the cycles being processed at a particular PL versus $t_\t{sp}$. Within some cycles being processed at PL3 many spikes occur simultaneously such that their processing $t_\t{sp}$ requires up to \SI{0.8}{ms}, where \SI{1}{ms} is the real-time constraint. Thus, the system is close to its performance limit. A conventional system without DVFS would have to be operated at PL3. In the DVFS approach only a little percentage of cycles are processed at higher PLs, thereby achieving nearly the energy efficiency of the low voltage operation at PL1.

\subsubsection{Bursting Network}
\label{sec:bursting_network}

\begin{figure}[htb]
	\centering
		\includegraphics[width=0.47\textwidth]{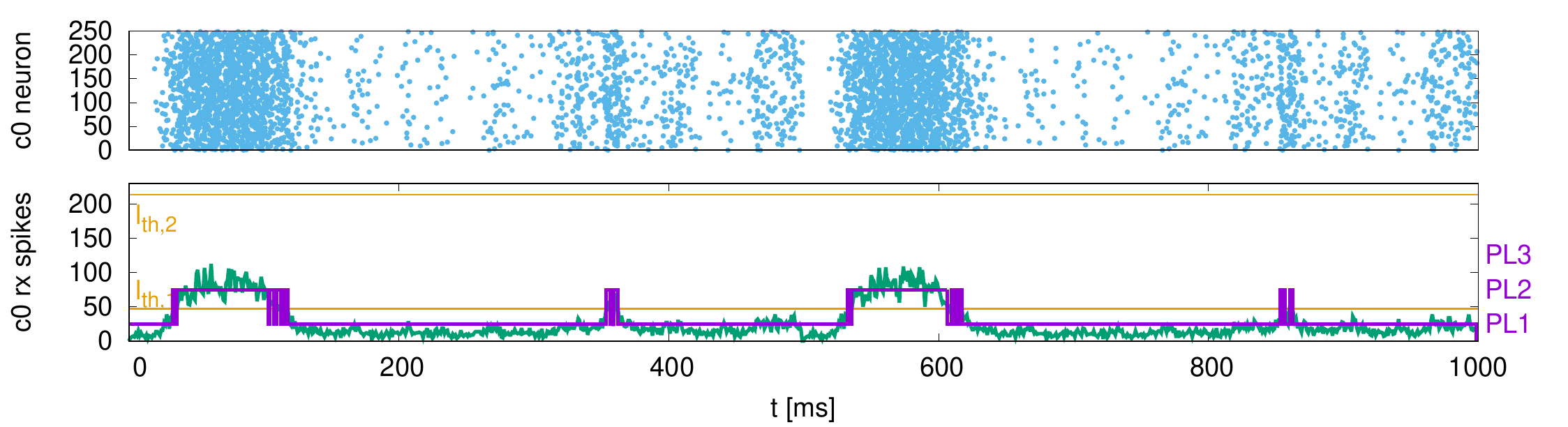}
	\caption{Bursting network benchmark spike train, sent spikes (blue), number of received spikes per core (green), PL selection thresholds (orange) and core PL (violet). Only core 0 is shown as activity on other cores is nearly identical.}
	\label{fig:burst_network_spikes_pl_v2}
\end{figure}


The second benchmark is a sparse random network generating network bursts, where the neurons collectively switch between UP states with high firing rate and DOWN states with low baseline activity. Such network events have been found both \emph{in-vitro} and \emph{in-vivo} at different spatial and temporal scales~\cite{Gigante2015}. The implemented model is based on~\cite{Mattia2012} and consists of $N=1000$ excitatory neurons recurrently connected with probability $p_\t{rec}=0.08$ (Fig.~\ref{fig:bursting_network}). The neurons are equally distributed over the 4 cores. Spike frequency adaptation (SFA) is implemented by creating an inhibitory synapse from each neuron to itself with a long synaptic decay time constant
. A background population (BG) of 200 Poisson neurons is connected with $p_\t{ext}=0.1$ to the excitatory population to generate a baseline firing activity in the network. Spikes from the BG population are stored before simulation in the DRAM and thus do not add significant workload.
Typical network dynamics are shown in Fig.~\ref{fig:burst_network_spikes_pl_v2}:
The network quickly enters an UP state with average firing rate higher than \SI{100}{\hertz} until the SFA silences the neurons into a DOWN state until the next network burst is initiated. The histogram of simulation cycles versus $t_\t{sp}$ is shown in Fig.~\ref{fig:pl_time_done_hist_burst}. Here the peak performance PL3 is not utilized.

\subsubsection{Asynchronous Irregular Firing Network}
\label{sec:ai_network}

\begin{figure}[htb]
	\centering
		\includegraphics[width=0.47\textwidth]{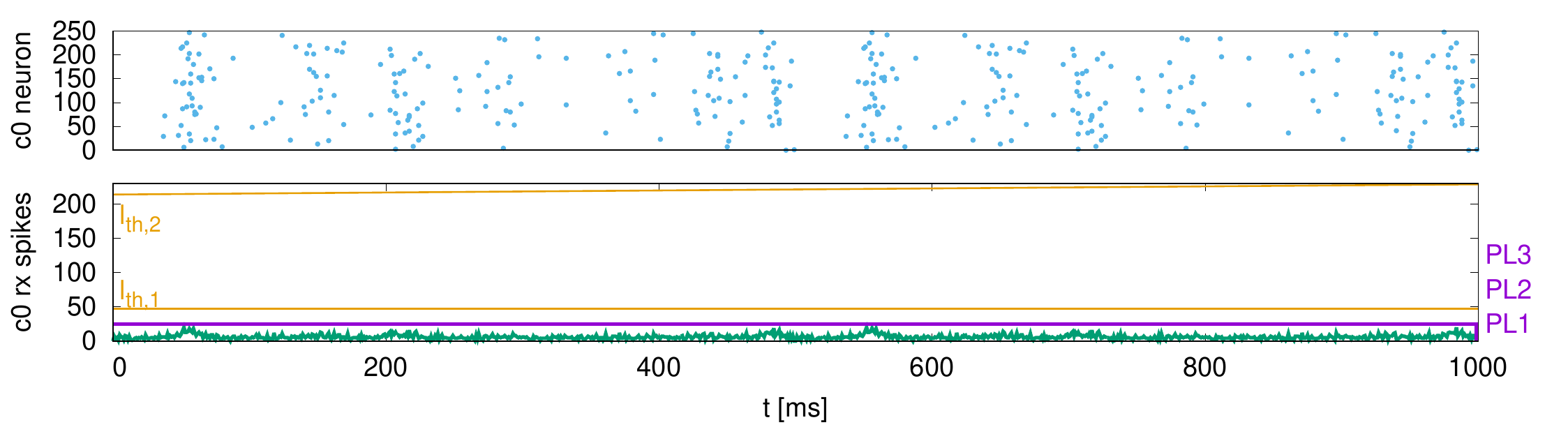}
	\caption{Asynchronous network benchmark spike train, sent spikes (blue), number of received spikes per core (green), PL selection thresholds (orange) and core PL (violet). Only core 0 is shown as activity on other cores is nearly identical.}
	\label{fig:asynchronous_network_spikes_pl_v2}
\end{figure}

\begin{figure*}[htb]
	\centering
		\subfigure[Synfire chain network                 \label{fig:pl_time_done_hist}]      {\includegraphics[width=0.30\textwidth]{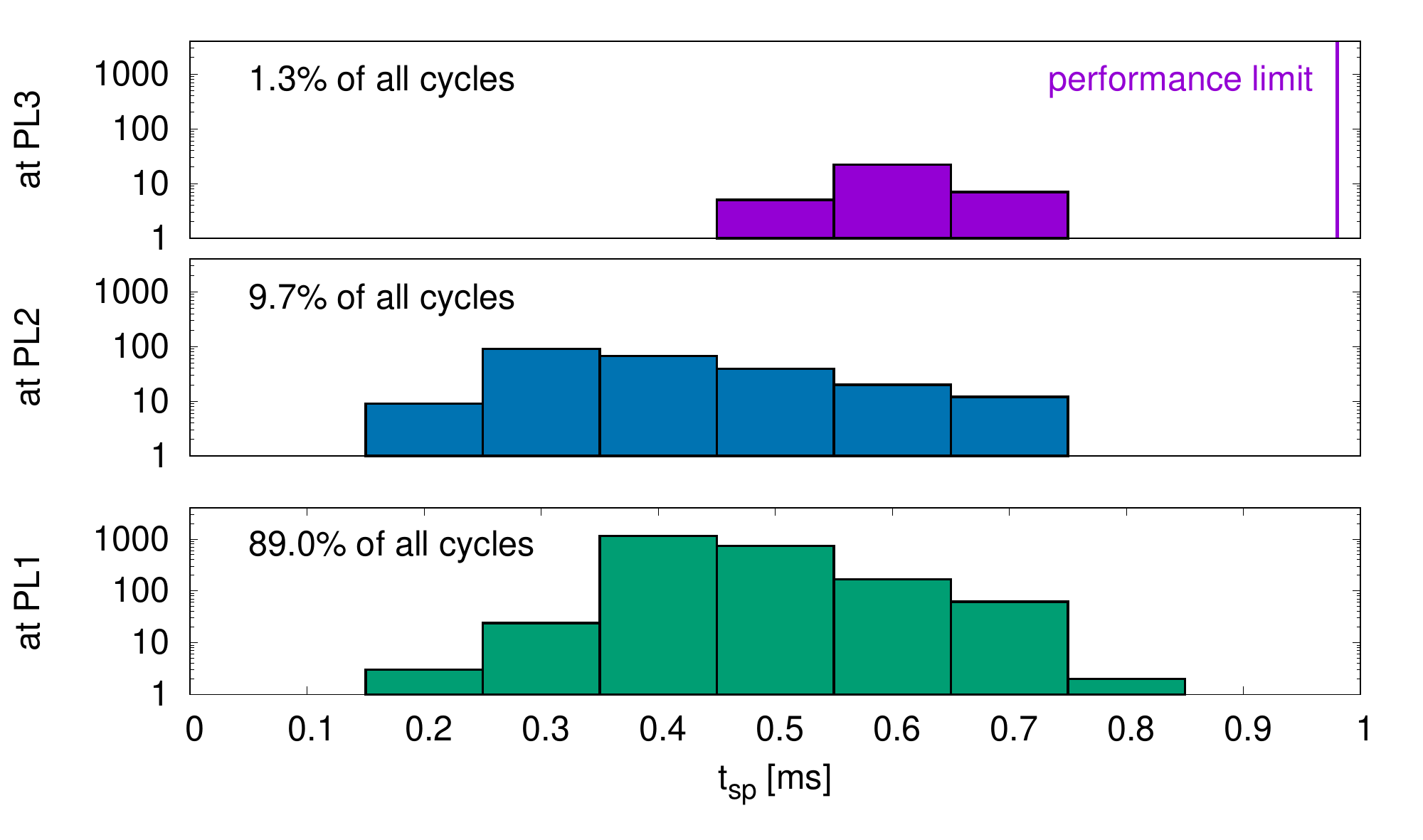}}
		\subfigure[Bursting network                      \label{fig:pl_time_done_hist_burst}]{\includegraphics[width=0.30\textwidth]{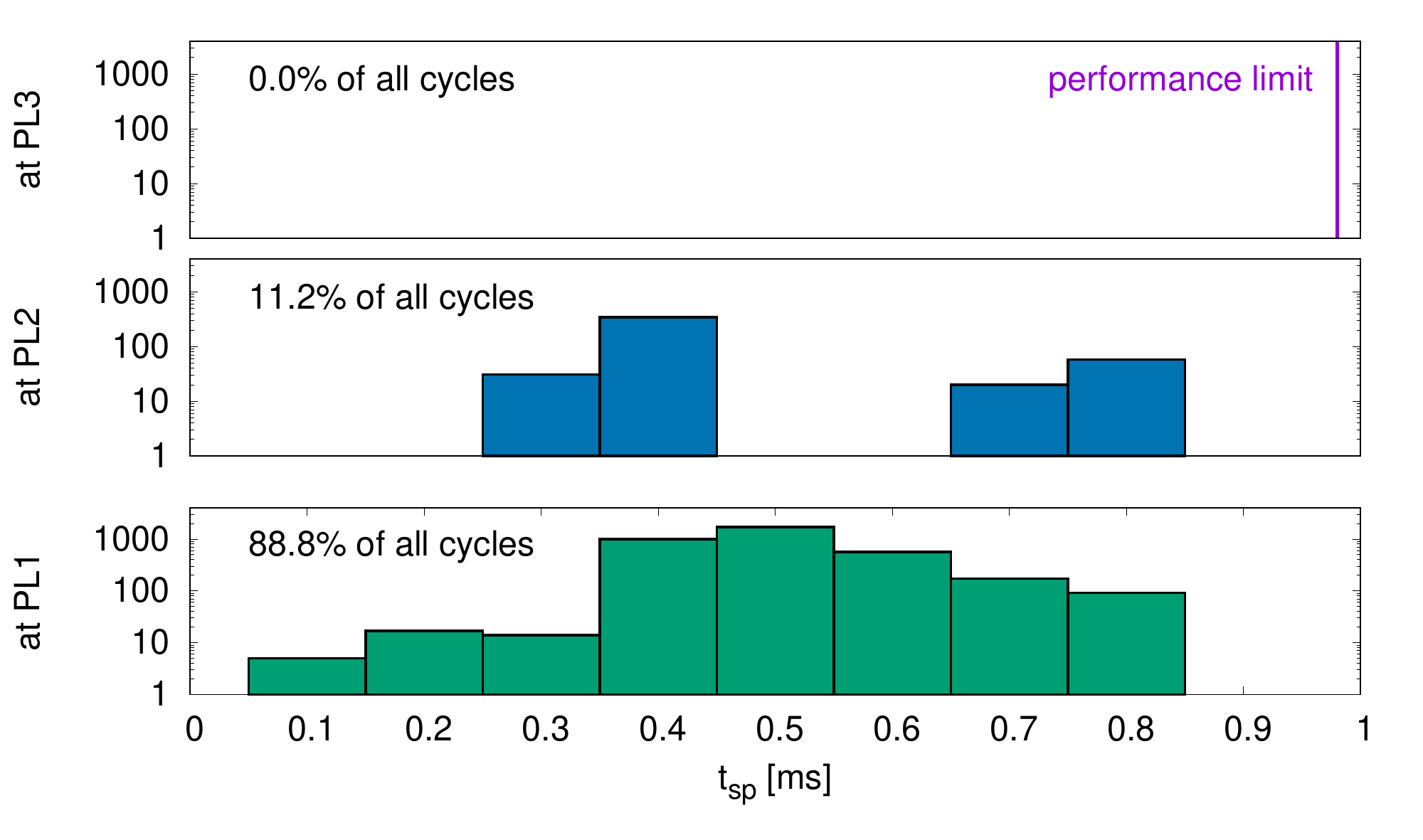}}
		\subfigure[Asynchronous irregular firing network \label{fig:pl_time_done_hist_async}]{\includegraphics[width=0.30\textwidth]{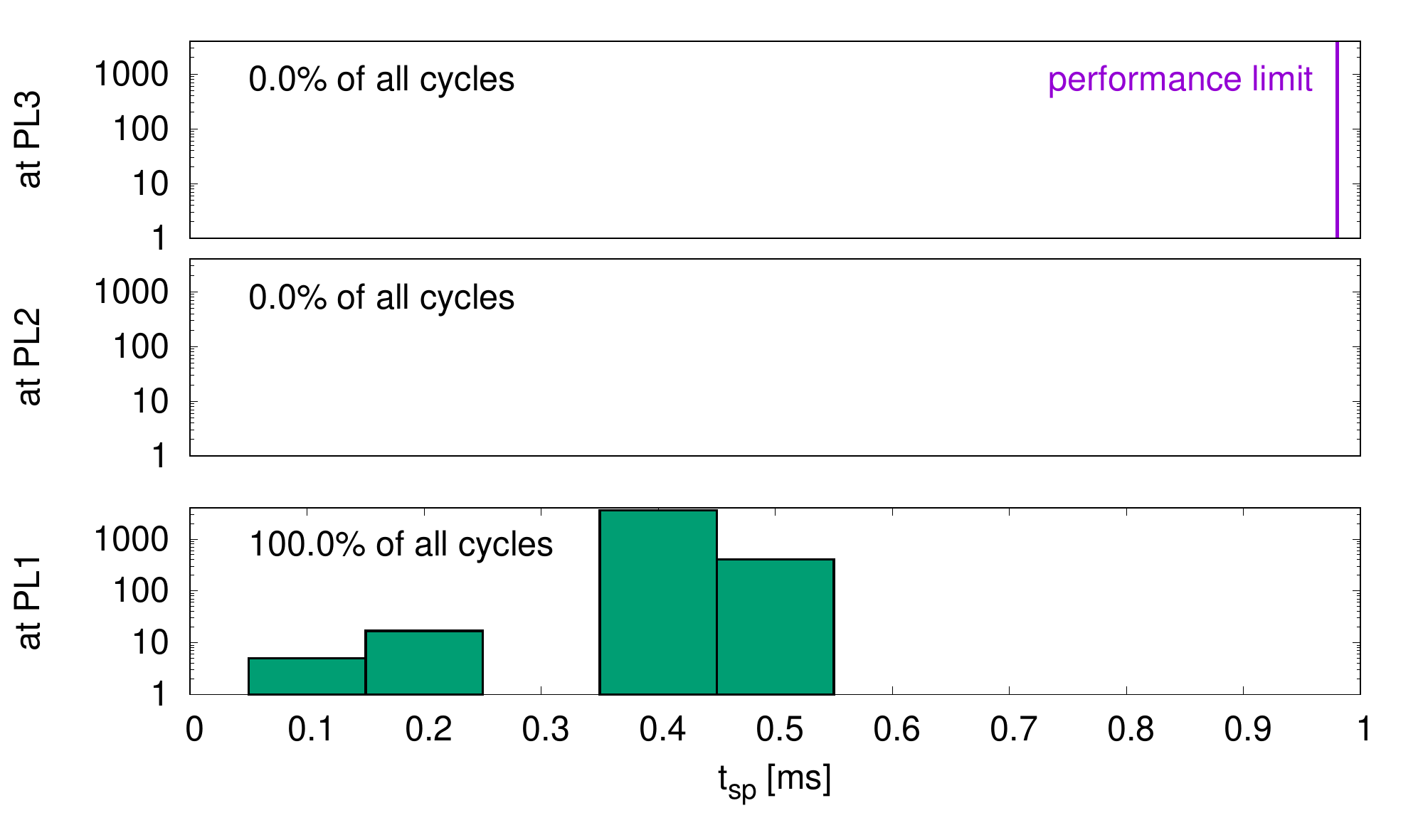}}
	\caption{Histograms of simulation cycles (\SI{1}{ms}) processed at different PLs versus $t_\t{sp}$}
	\label{fig:pl_time_done_hist_all}
\end{figure*}

A sparse random network with asynchronous irregular firing activity serves as the third benchmark. The limited size of the prototype system does not allow to implement the standard benchmark for sparse random networks~\cite{Vogels2005}, which is commonly used to benchmark SNN simulators~\cite{Brette2007,Zenke2014} or neuromorphic hardware~\cite{Knight2016}. Instead, we use the same network architecture as for the bursting network (Fig.~\ref{fig:ai_network}) and disable the spike frequency adaption. Additionally, the recurrent connection probability is reduced to $p_\t{rec}=0.02$ such that the network stays in a low-rate asynchronous irregular firing regime, as can be seen in Fig.~\ref{fig:asynchronous_network_spikes_pl_v2}. Fig.~\ref{fig:pl_time_done_hist_async} shows the corresponding histogram of simulations cycles versus $t_\t{sp}$. In this case only the lowest PL is required. The system automatically remains in its most energy efficient operation mode.

\subsection{Power Management Results}
To assess the benefit of the dynamic power management, we compare the power consumption for the benchmarks at the highest performance level (PL3) and using DVFS.
The power measurement is done differentially. First the total power \(P_{0}\) is measured with the neuromorphic experiment running on the chip. Then spike sending is deactivated and the power \(P_{1}\) is measured. Since no spike is sent, no spike is received and processed. The difference \(P_{\t{syn}}=P_{0}-P_{1}\) is the synapse processing power. Then all neural processing is deactivated and upon $t_\t{sys}$ timer interrupt only empty interrupt handlers are called. The power at this stage is measured as \(P_{2}\). \(P_{\t{neur}}=P_{1}-P_{2}\) is the neuron processing power. Then the ARM cores are deactivated and the power is measured as \(P_{3}\). \(P_{\t{BL}}=P_{2}-P_{3}\) is the baseline power. When DVFS is enabled, the PL is determined by the network activity. Thus, after measuring \(P_{0}\) with the complete software autonomously switching PLs during simulation, the percentage of simulation time at the \SI{3}{PLs} are recorded and then applied to the simulations when measuring \(P_{1}\) and \(P_{2}\).

\unsetrowcolors{}
\begin{table*}[htb]
  \begin{center}
  \begin{minipage}{0.78\textwidth}
    \centering
    \renewcommand{\arraystretch}{1.1}
    \caption{Benchmark power results}\label{tab:powermeas}
    \centering
    \footnotesize
    \begin{tabular}{ll@{\extracolsep{\fill}}cccccccc} \toprule
                                                                                                                       &                                                        & \multicolumn{2}{c}{local}     & \multicolumn{2}{c}{synfire}   & \multicolumn{2}{c}{bursting} &  \multicolumn{2}{c}{async}              \\
                                                                                                                       &                                                        & only PL3                      & only PL1                      & only PL3                     &  DVFS                                   &  only PL3 & DVFS & only PL3 &  DVFS          \\ \midrule \addlinespace
      \cellcolor{tablecol-odd}                                                                                         & total\tablenote{excluding unused test chip components} & 138.2                         & 72.7                          & 135.6                        &  71.2                                   &  136.5    & 71.6 & 133.8    &  67.3          \\ \addlinespace \rowcolor{tablecol-even}
      \cellcolor{tablecol-odd}                                                                                         & infrastructure\tablenote{timer, router, LPDDR2}        & 48.2                          & 48.2                          & 48.2                         &  48.2                                   &  48.2     & 48.2 & 48.2     &  48.2          \\
      \cellcolor{tablecol-odd}                                                                                         & baseline                                               & 71.2                          & 14.9                          & 76.2                         &  17.0                                   &  76.0     & 17.0 & 76.4     &  14.7          \\ \rowcolor{tablecol-even}
      \cellcolor{tablecol-odd}                                                                                         & neuron                                                 & 3.0                           & 1.7                           & 7.7                          &  3.6                                    &  7.8      & 3.5  & 7.5      &  3.6           \\
      \cellcolor{tablecol-odd}                                                                                         & synapse                                                & 15.9                          & 7.9                           & 3.5                          &  2.4                                    &  4.5      & 2.9  & 1.7      &  0.8           \\ \addlinespace \rowcolor{tablecol-even}
      \cellcolor{tablecol-odd}                                 \multirow{-7}{*}{\rotatebox{90}{\textbf{power [mW]}}} & PE                                                     & 90.0                          & 24.5                          & 87.4                         &  23.0                                   &  88.3     & 23.4 & 85.6     &  19.1          \\ \addlinespace \midrule \addlinespace
      \multicolumn{2}{l}{SynEvents/s}                                                                                  & \multicolumn{2}{c}{16,000,000}                         & \multicolumn{2}{c}{3,030,000} & \multicolumn{2}{c}{2,240,000} & \multicolumn{2}{c}{489,000}  \\ \addlinespace \rowcolor{tablecol-even}
      \multicolumn{2}{l}{E/SynEvent\tablenote{total power divided by SynEvents/s} [nJ]}                                & 8.6                                                    & 4.5                           & 44.7                          & 23.5                         &  61.0                                   &  32.0     & 274  & 138      \\
      \multicolumn{2}{l}{E/SynEvent\tablenote{PE power divided by SynEvents/s} (PE only) [nJ]}                         & 5.6                                                    & 1.5                           & 28.8                          & 7.59                         &  39.4                                   &  10.5     & 175  & 39.1     \\ \addlinespace
      \bottomrule
    \end{tabular}
  \end{minipage}
  \end{center}
\end{table*}

Tab.~\ref{tab:powermeas} summarizes the power measurement results of the system for the benchmarks. For comparison, we also include the locally-connected network used for the power model parameter extraction in Sec.~\ref{sec:pmextraction}. The testchip as shown in Fig.~\ref{fig:Santos28_Blockdiagram} contains only 4 PEs for prototyping purposes, for which the relative impact of infrastructure power, including the LPDDR2 memory interface is relatively high. It is expected that for future neuromorphic SoCs the infrastructure overhead is somehow balanced with respect to the number and performance of the PEs on the chip. Thus, the further comparison of energy efficiency is focused on the PEs only, which benefit from the proposed DVFS technique. Using DVFS, baseline power can be reduced by $\approx \SI{80}{\%}$, neuron power by up to $\approx \SI{50}{\%}$ and synapse power by between $\approx \SI{35}{\%}$ and $\approx \SI{50}{\%}$, depending on the experiment. The total PE power reduction by means of DVFS is $\approx \SI{73}{\%}$. For comparison with other systems, we also calculate the energy per synaptic event (total energy vs.~total synaptic events), which reaches its minimum value of \SI{1.5}{nJ} PE energy at highest utilization within the locally connected network, similar to the benchmark used in~\cite{Stromatias2013}.


\subsection{Results Comparison}
\label{sec:resultscomparison}

\def\tbrefNeurogrid{\cite{Benjamin2014}}
\def\tbrefROLLS{\cite{qiao2015reconfigurable,Indiveri2015}}
\def\tbrefHiAER{\cite{Yu2012}}
\def\tbrefcxQuad{\cite{moradi2017scalable,Indiveri2015} \tablenote{Synaptic input event of \SI{0.134}{fJ} is broadcast to all \SI{256}{neurons} per core. Total power assumes \SI{1024}{neurons} firing at \SI{30}{Hz} connected to \SI{256}{targets} each consuming \SI{360}{uW} at \SI{1.3}{V}.}}
\def\tbrefBrainScales{\cite{Schmitt2017}}
\def\tbrefTitan{\cite{mayr2016,noack14b} \tablenote{Assuming accelerated operation with speed-up factor of 100, power draw of \SI{15}{mW}, 128 inputs firing at \SI{1}{kHz} connected to \SI{64}{targets} each.}}
\def\tbrefTrueNorth{\cite{merolla2014million}}
\def\tbrefOdin{\cite{frenkel2018}}
\def\tbrefLoihi{\cite{Davies2018}}
\def\tbrefmemristor{\cite{du15}}
\def\tbrefSpinnaker{\cite{Stromatias2013}\tablenote{Results extracted from 1st column of Table~III in~\cite{Stromatias2013}, infrastructure power assumed as \SI{11}{W} per board, cf. Figure~6 of~\cite{Stromatias2013}}}
\def\tbrefthiswork{this work\tablenote{at PL1 \SI{0.7}{V}}}

\textcolor{red}{Although this work is focused on a dynamic power management technique for event-based digital neuromorphic systems, thereby not being directly comparable to other neuromorphic approaches, Tab.~\ref{tab:compare} compares the achieved energy consumptions also of different neuromorphic systems. 
For realistic comparison two metrics for synaptic energy are considered. First, the energy per synaptic event as in total energy divided by synaptic events processed across the system and second the incremental energy for one more synaptic event. These metrics are not scaled to the semiconductor technology, since completely different circuit approaches provide their optimal results in different technology nodes. For a fair comparison these metrics would have to be extracted from the same benchmark running on the different neuromorphic hardware systems \cite{diamond2016comparing}. 
We did not put memristor systems in the table, as the approach seems too different and no large-scale memristor systems have been reported. However, for comparison we would still name a few figures for memristor synapse arrays: Du et al.~\cite{du15} report measurements of \SI{4.7}{pJ} for potentiation/depression at a single synapse, but without the circuit overhead. Chakma et al.~\cite{chakma2018memristive} simulate memristors and CMOS neurons, reporting between 0.1 and \SI{10}{pJ/synaptic operation} for both learning and passive memristors, with additional \SI{10}{pJ/spike} for the CMOS neurons. Analog subthreshold systems \cite{Benjamin2014,qiao2015reconfigurable,Yu2012,moradi2017scalable} and mixed-signal systems \cite{Schmitt2017,mayr2016,noack14b} use analog circuits to mimic neural and synaptic behavior. Custom-digital neuromorphic chips \cite{merolla2014million,frenkel2018,Davies2018} use event-based processing in custom non-processor units. They can be implemented in nanometer CMOS technologies and show low energy per synaptic event regarding total power in the same order of magnitude as the analog and mixed-signal approaches.} 
\textcolor{red}{Multi-processor-based neuromorphic systems, such as this work, trade off much higher system flexibility due to software defined neuromorphic processing by two to three orders of magnitude higher energy consumption. However, the scaling of semiconductor technology together with dynamic power management techniques such as the proposed DVFS in this work reduces this gap. Compared to SpiNNaker approximately $10\t{x}$ reduction of neuron processing energy and PE energy per synaptic event is achieved. For our benchmarks on this work, as summarized in Tab.~\ref{tab:powermeas}, the application of DVFS results in 73\% total PE power reduction. The proposed technique is also applicable to custom-digital neuromorphic systems which operate in an event-driven fashion.}

\setrowcolors{}
\begin{table*}[htb]
  \begin{minipage}{0.97\textwidth}
    \centering
    \renewcommand{\arraystretch}{1.1}
    \caption{Efficiency comparison of neuromorphic hardware systems}\label{tab:compare}
    \centering
    \footnotesize
    \begin{tabular}{llcccp{2.5cm}c} \toprule \setrowcolors{}
       System               & Reference           & Type                     & Techn.\ node & Neuron power      & E/SynEvent                                                                         & E/additional SynEvent \\
                            &                     &                          & [\si{nm}]    & [\si{nJ}/\si{ms}] & [\si{pJ}]                                                                          & [\si{pJ}]             \\ \midrule
       Neurogrid            & \tbrefNeurogrid{}   & analog sub-Vt            & 180          &                   & 119                                                                                & n.a.                  \\ 
       ROLLS                & \tbrefROLLS{}       & analog sub-Vt            & 180          &                   & n.a.                                                                               & 0.077                 \\ 
       HiAER-IFAT           & \tbrefHiAER{}       & analog sub-Vt            & 130          &                   & 50                                                                                 & n.a.                  \\ \midrule
       cxQuad               & \tbrefcxQuad{}      & mixed-signal             & 180          &                   & 46                                                                                 & 0.134                 \\ 
       BrainScales          & \tbrefBrainScales{} & mixed-signal accelerated & 180          &                   & 100                                                                                & n.a.                  \\ 
       Titan                & \tbrefTitan{}       & mixed-signal             & 28           &                   & 18                                                                                 & n.a.                  \\ \midrule
       TrueNorth            & \tbrefTrueNorth{}   & custom digital           & 28           &                   & 26                                                                                 & n.a.                  \\ 
       Odin                 & \tbrefOdin{}        & custom digital           & 28           &                   & n.a.                                                                               & 9.8                   \\ 
       Loihi                & \tbrefLoihi{}       & custom digital           & 14           & 0.052             & n.a.                                                                               & 23.6                  \\ \midrule 

       SpiNNaker            & \tbrefSpinnaker{}   & MPSoC                    & 130          & 26                & 13300\tablenote{regarding PE power only} (19300) \tablenote{regarding total power} & 8000                  \\

       SpiNNaker2 prototype & \tbrefthiswork{}    & MPSoC                    & 28           & 2.19              & 1500 (4500)                                           & 450                   \\

      \bottomrule
    \end{tabular}
    \vspace{0.8ex}

    For all systems we report the best case with highest energy efficiency found in literature.  Note that results are not fully commensurable, especially for the metric considering total power, as different benchmarks are applied.
  \end{minipage}
\end{table*}



%% file: dvfs_architecture_exploration.tex
\begin{figure}[htb]
	\centering
		\includegraphics[width=0.47\textwidth]{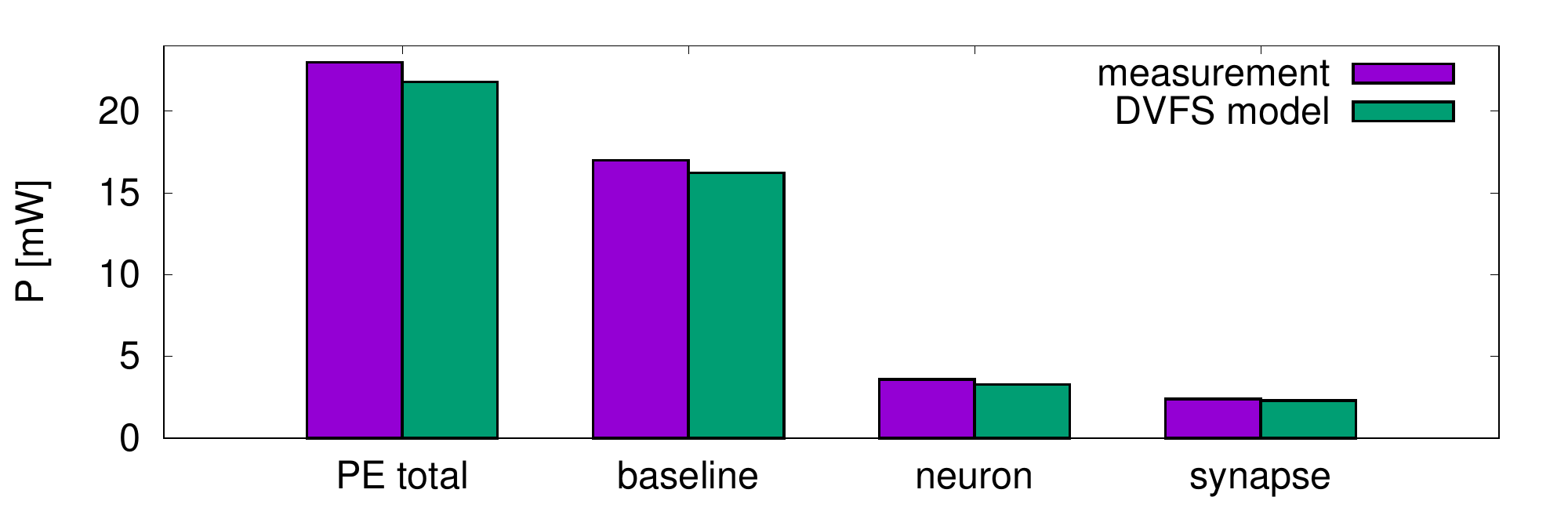}
	\caption{Measured power consumption versus DVFS model (synfire chain)}\label{fig:dvfs_model_vs_meas}
\end{figure}

Based on a numerical power management model from Sec.~\ref{sec:pmtheory} and the extracted model parameters from the \SI{28}{nm} test chip, an exploration of the DVFS architecture is performed. The synfire chain from Sec.~\ref{sec:synfire_chain} is chosen as benchmark, since it shows highly dynamic workload with the temporary demand for peak performance at the highest PL.
Fig.~\ref{fig:dvfs_model_vs_meas} shows the measured PE power consumption compared to the DVFS model parameters, extracted with a fanout of 80 similar to the synfire chain configuration as shown in Sec.~\ref{sec:pmextraction}. The power values match with acceptable accuracy.  

\begin{figure}[htb]
	\centering
		\includegraphics[width=0.47\textwidth]{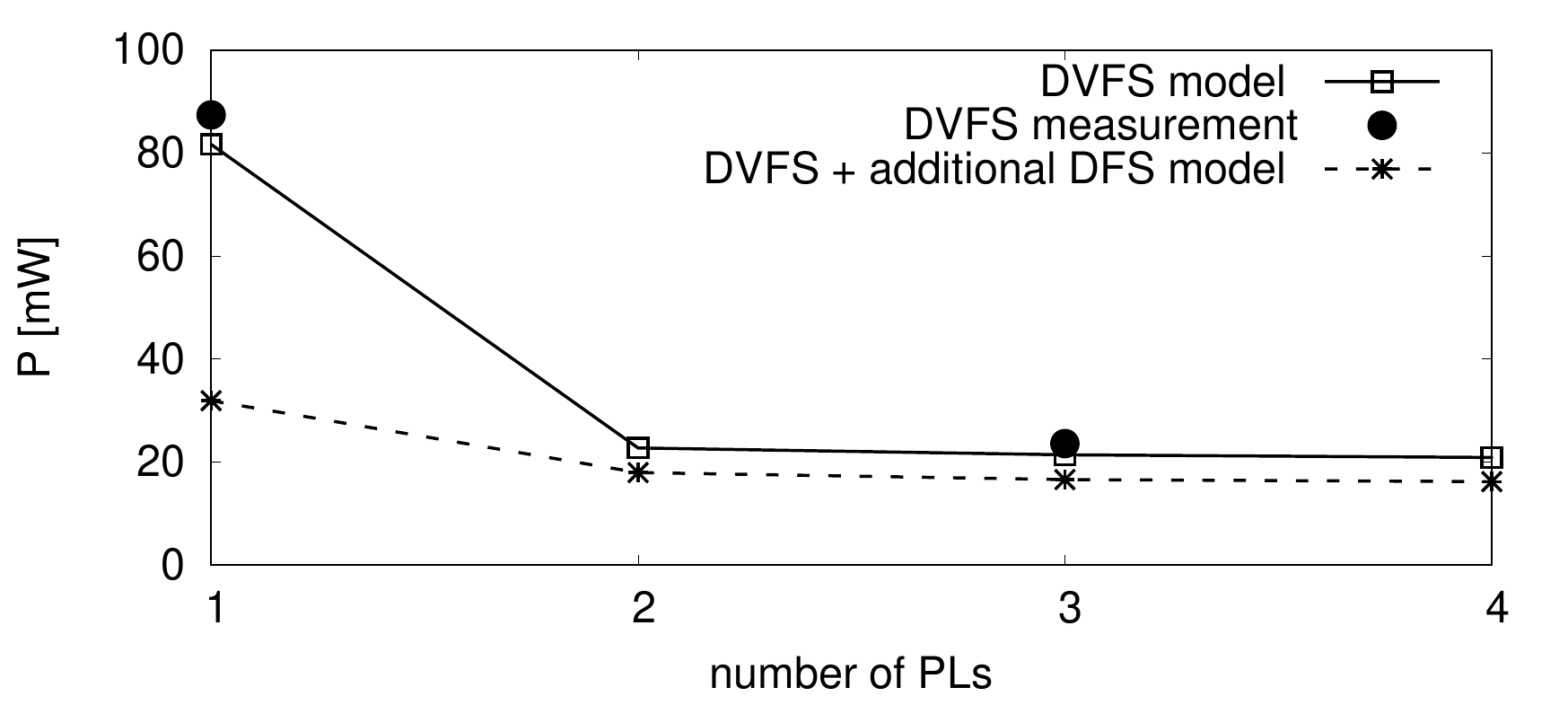}
	\caption{\textcolor{red}{PE power prediction (baseline power, neuron power, synapse power) of synfire chain benchmark for different number of PLs and one scenario with an additional DFS of \SI{10}{MHz} level at the lowest available PL}}\label{fig:npm_model_pl_levels_power}
\end{figure}

For DVFS architecture exploration the number of PLs with dedicated supply rails is considered as parameter, since the hardware overhead of separated supply rails, power switches and external voltage regulators is the main system overhead of the DVFS approach. Therefore, the power saving potential versus the number of PLs is analyzed. It is based on the workload threshold method described in this paper, based on \SI{3}{PLs}. If the number of PLs is lower, e.g. 2, it is directly switched to PL3 when $l_{\t{th},1}$ is achieved. PL2 is omitted in this case. Fig.~\ref{fig:npm_model_pl_levels_power} shows the simulated and measured PE power for different numbers of PLs. They always include the \SI{500}{MHz} at \SI{1.0}{V} PL, since this one provides the required peak performance. It can be seen that already with \SI{2}{PLs} the PE power can be reduced by \SI{70}{\%}. Addition of a third level results in \SI{73}{\%} power reduction. \textcolor{red}{A hypothetical 4th PL has been added to a scheme of (0.70V, 0.80V, 0.90V, 1.0V) with (125MHz, 300MHz, 400MHz, 500MHz) and analyzed using the model. Adding this results in only  \SI{1}{\%} additional power reduction compared to three PLs.} From this it is concluded that more than \SI{2}{PLs} with distinct supply rails do not gain much additional efficiency, justifying their additional overhead.  

Without the insertion of an additional supply rail, a dynamic frequency scaling (DFS) performance level can be added.
This is based on the supply rail selection of the lowest PL but has a much lower ADPLL clock generator frequency setting.
The PE can switch to this DFS level, after the neuron and synapse computation within a simulation time step is done after $t_\t{sp}$.
This allows the reduction of baseline power after the computation is done, while still allowing the core to react on interrupts.
Fig.~\ref{fig:npm_model_pl_levels_power} shows an example DFS analysis result where an additional DFS level with \SI{10}{MHz} clock frequency is assumed. This reduces the total PE baseline power after $t_\t{sp}$ close to the leakage power value of $P_{BL,\t{leak},1}=8.94\t{mW}$, $P_{BL,\t{leak},2}=20.03\t{mW}$ and $P_{BL,\t{leak},3}=28.53\t{mW}$, respectively.
Due to the high baseline power portion in this particular implementation, this results in \SI{62}{\%} power reduction at the highest PL.
Adding one or two more PLs, power can be reduced by additionally \SI{45}{\%} and \SI{49}{\%}, respectively.

%% file: conclusion.tex
A DVFS power management approach for event-based neuromorphic real-time simulations on MPSoCs has been presented. Its effectiveness has been demonstrated with a \SI{28}{nm} CMOS prototype. For a neuromorphic benchmark application, the PE power including baseline power and energy consumption for neuromorphic processing can be significantly reduced by up to $\approx \SI{73}{\%}$ compared to non-DVFS operation while maintaining biological real-time operation. 
Using the presented neuromorphic power management model, energy consumption of the next generation large scale neuromorphic many core systems can be estimated. It helps to design both the power management hardware architecture, the software flow and the strategy for mapping a neuromorphic problem to the system with energy awareness. The results will directly flow into the SpiNNaker2 neuromorphic many core system, which is currently under development.


%% file: bio.tex
\begin{IEEEbiography}[{\includegraphics[width=1in,height=1.25in,clip,keepaspectratio]{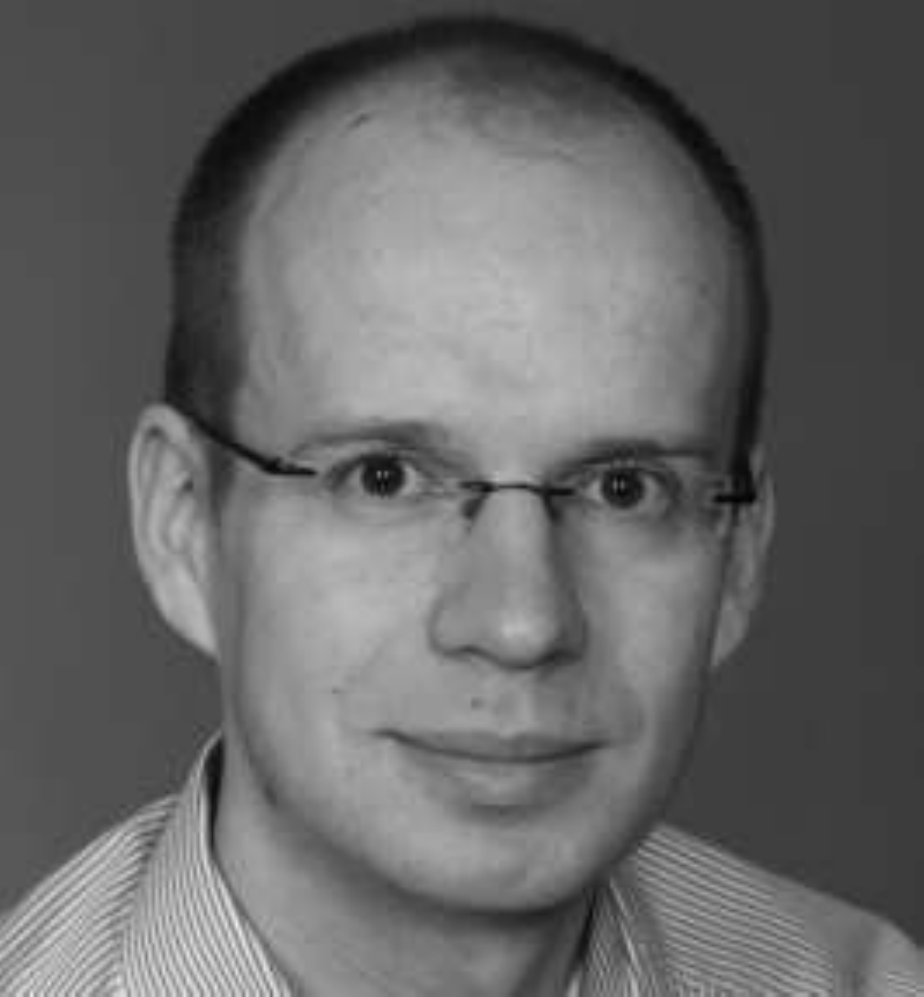}}]{Sebastian Höppner}
is a Research Group Leader and Lecturer with the Chair of Highly-Parallel VLSI-Systems and Neuromorphic Circuits. He received the Dipl.-Ing. (M.Sc.) in Electrical Engineering in 2008 and his Ph.D. in 2013 (received Barkhausen Award), both Technische Universität Dresden, Germany. His research interests include circuits for low-power systems-on-chip in advanced technology nodes, with special focus on clocking, data transmission and power management. He has experience in designing full-custom circuits for multi-processor systems-on-chip (MPSoCs), like ADPLLs, register files and high-speed on-chip and off-chip links, in academic and industrial research projects. He has been managing the full-custom circuit design and SoC integration for more than 12 MPSoC chips in 65nm, 28nm and 22nm CMOS technology. Currently he leads the chip design of the SpiNNaker2 neuromorphic computing system within the Human Brain Project(HBP). He is author or co-author of more than 56 publications and 10 patents (5 issued, 5 pending) in the above fields.
\end{IEEEbiography}

\begin{IEEEbiography}[{\includegraphics[width=1in,height=1.25in,clip,keepaspectratio]{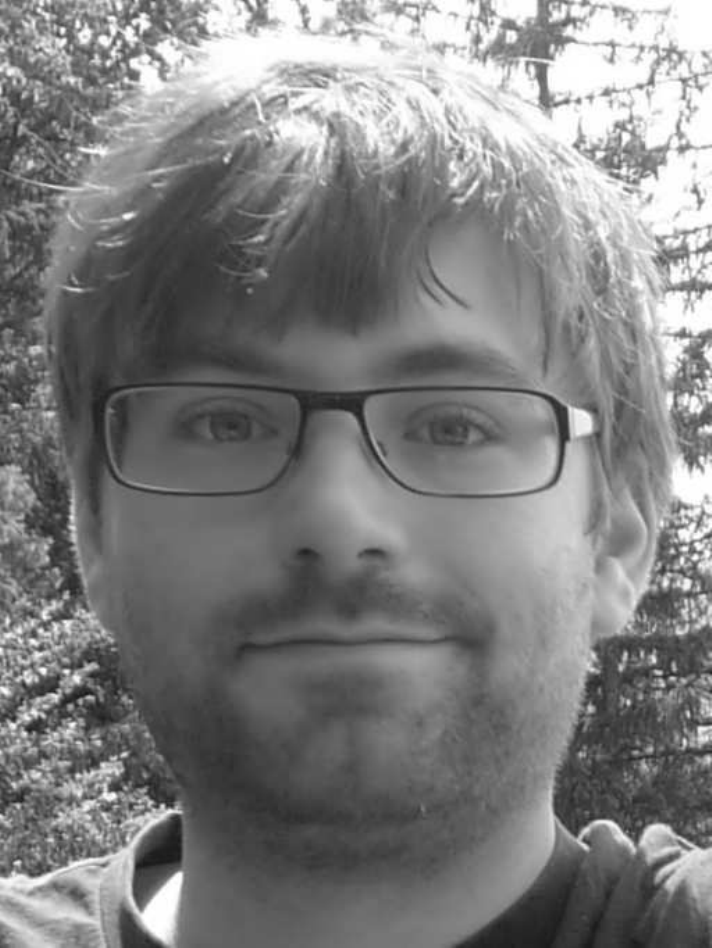}}]{Bernhard Vogginger}
received the diploma in physics from the University of Heidelberg, Heidelberg, Germany, in 2010. Currently, he is a research associate at the Chair of Highly-Parallel VLSI-Systems and Neuromorphic Circuits at Technische Universität Dresden, Germany, where he is pursuing a PhD under the supervision of Prof. Christian Mayr. His research interests include neuromorphic engineering, neural computation and deep learning.
\end{IEEEbiography}

\begin{IEEEbiography}[{\includegraphics[width=1in,height=1.25in,clip,keepaspectratio]{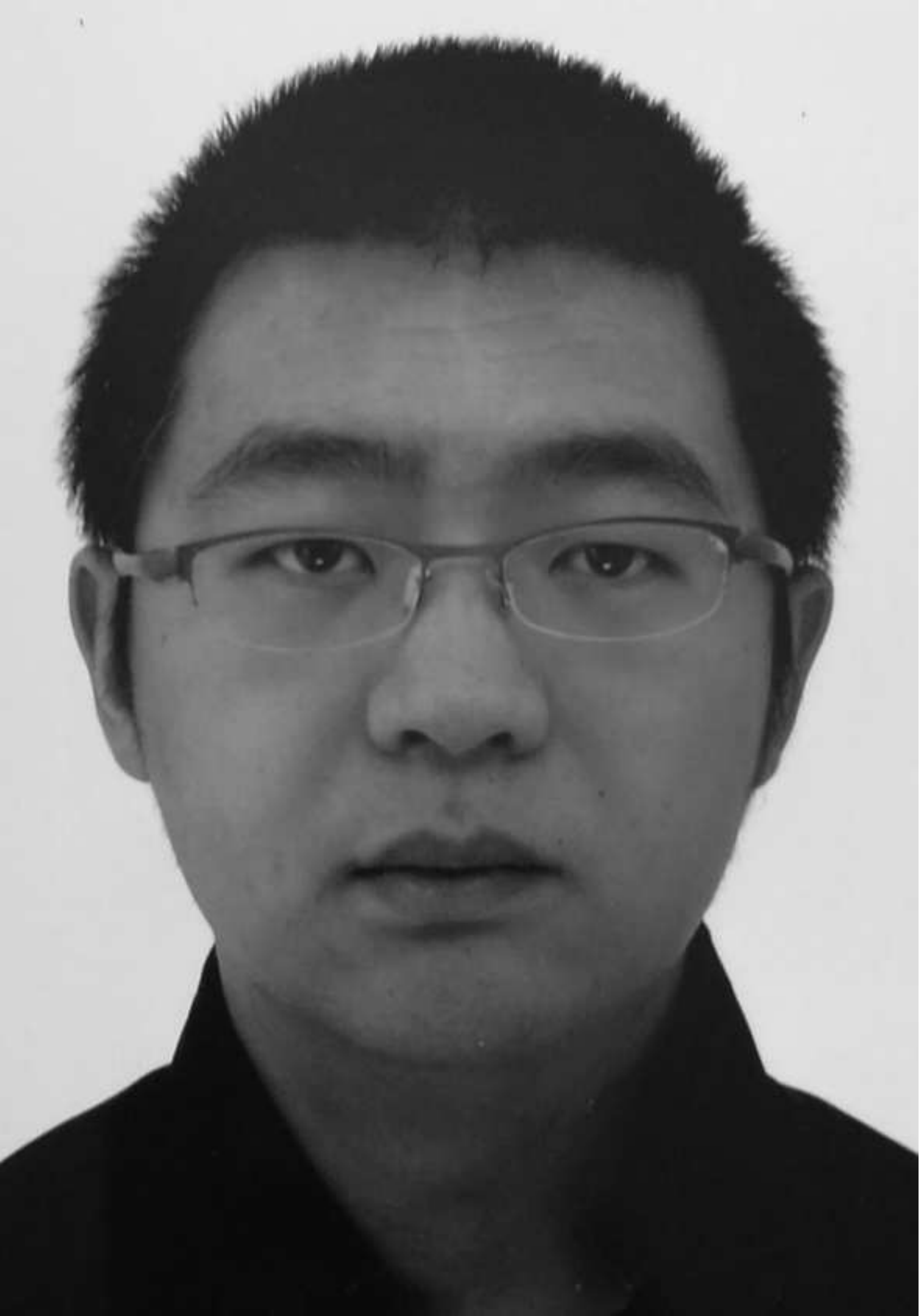}}]{Yexin Yan}
received the Dipl.-Ing. (M.Sc.) in Electrical Engineering from Technische Universität Dresden, Germany, in 2016. He is currently pursuing the Ph.D. at the Chair of Highly-Parallel VLSI-Systems and Neuromorphic Circuits at Technische Universität Dresden. His research interests include hardware-software co-design for low power neuromorphic applications.
\end{IEEEbiography}

\begin{IEEEbiography}[{\includegraphics[width=1in,height=1.25in,clip,keepaspectratio]{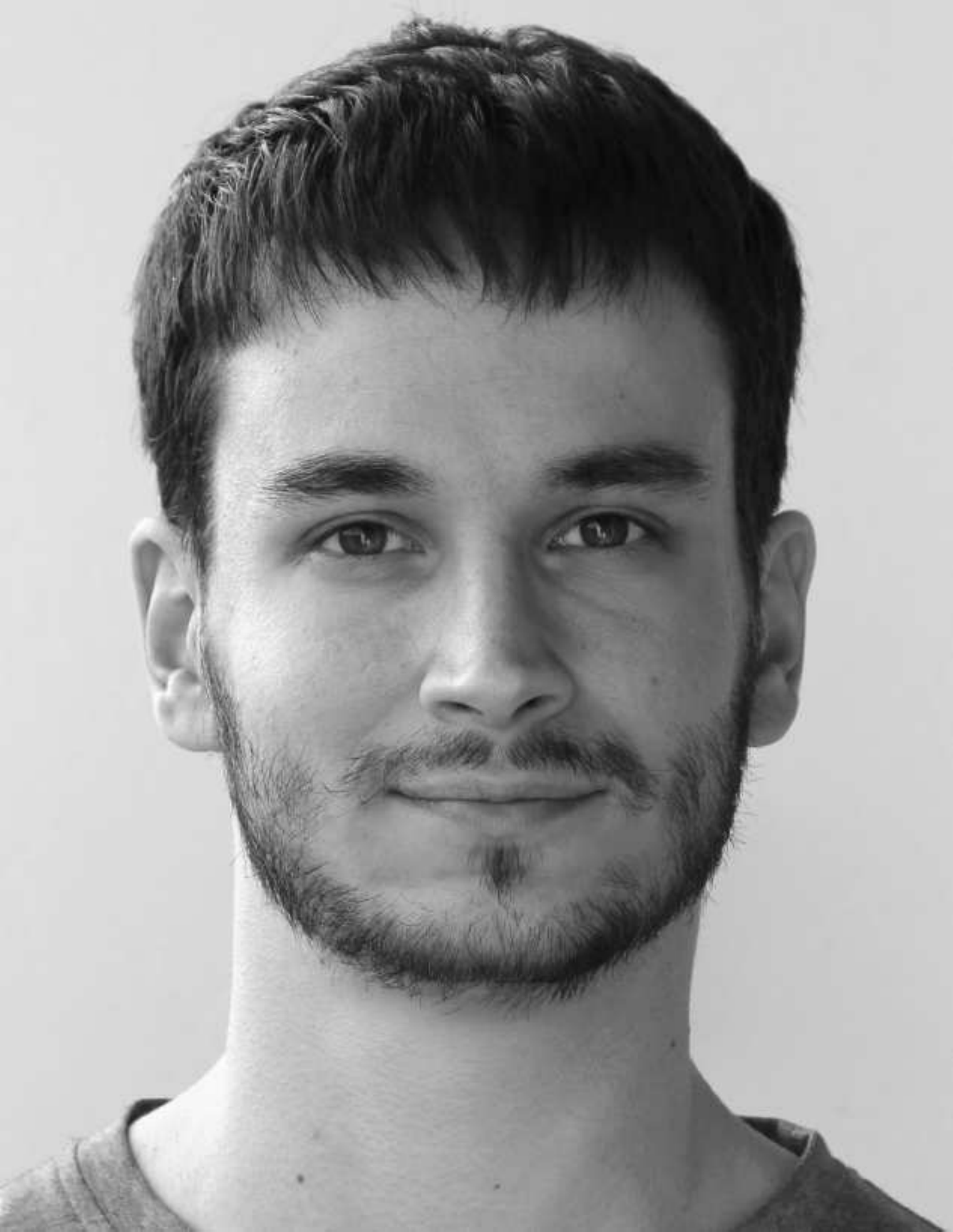}}]{Andreas Dixius}
received the Dipl.-Ing. (M.Sc.) in Information Systems Engineering at Technische Universität Dresden, Germany in 2014. Since 2014 he has been a research assistant at the Chair of Highly-Parallel VLSI-Systems and Neuromorphic Circuits, Technische Universität Dresden, Germany. His research interests include on-chip timing-detection.
\end{IEEEbiography}

\begin{IEEEbiography}[{\includegraphics[width=1in,height=1.25in,clip,keepaspectratio]{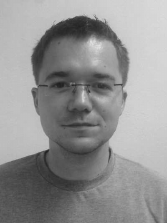}}]{Stefan Scholze}
received the Dipl.-Ing. (M.Sc.) in Information Systems Engineering from Technische Universität Dresden, Germany in 2007. Since 2007, he has been a research assistant at the Chair of Highly-Parallel VLSI-Systems and Neuromorphic Circuits, Technische Universität Dresden, Germany. His research interests include design and implementation of low-latency communication channels and systems.
\end{IEEEbiography}

\begin{IEEEbiography}[{\includegraphics[width=1in,height=1.25in,clip,keepaspectratio]{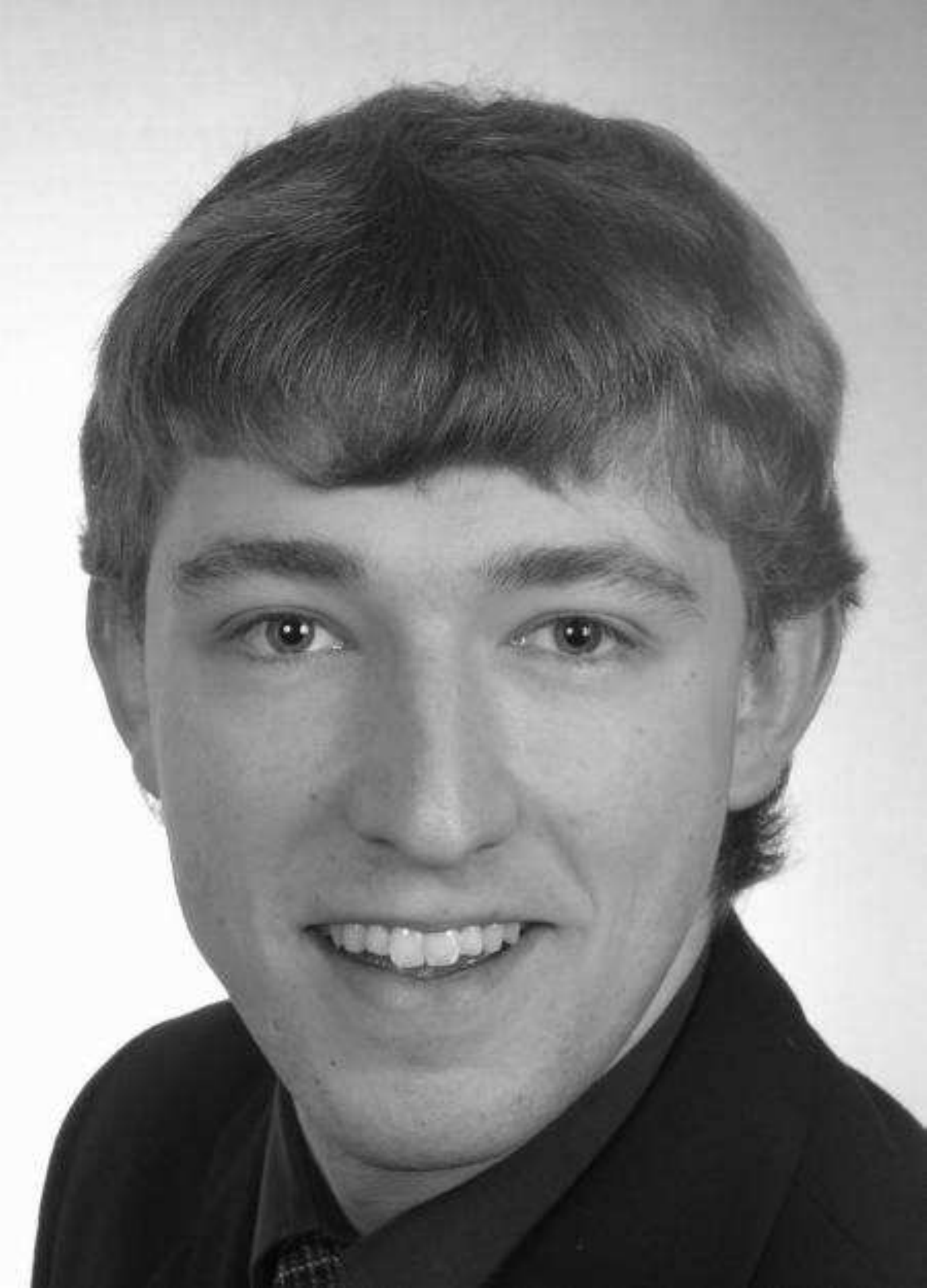}}]{Johannes Partzsch}
obtained his M.Sc.\  in Electrical Engineering in 2007 and his PhD in 2014, both from Technische Universität Dresden. He is currently a Research Group Leader at the Chair of Highly-Parallel VLSI-Systems and Neuromorphic Circuits, Technische Universität Dresden, Germany. His research interests include neuromorphic systems design, topological analysis of neural networks and technical application of bio-inspired systems. He is author or co-author of more than 45 publications.
\end{IEEEbiography}

\begin{IEEEbiography}[{\includegraphics[width=1in,height=1.25in,clip,keepaspectratio]{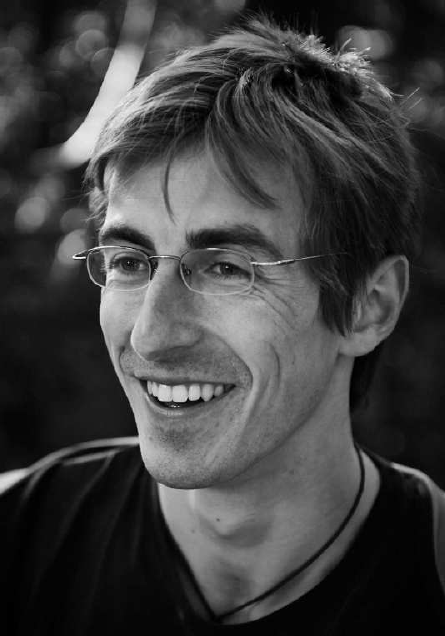}}]{Felix Neumärker}
received the Dipl.-Ing. (M.Sc.) in Electrical Engineering from Technische Universität Dresden, Germany, in 2015.
He is currently working as research associate with the Chair of Highly-Parallel VLSI-Systems and Neuromorphic Circuits at Technische Universität Dresden.
His research interests include software and circuit design and for MPSoCs with special focus on neuromorphic computing.
\end{IEEEbiography}

\begin{IEEEbiography}[{\includegraphics[width=1in,height=1.25in,clip,keepaspectratio]{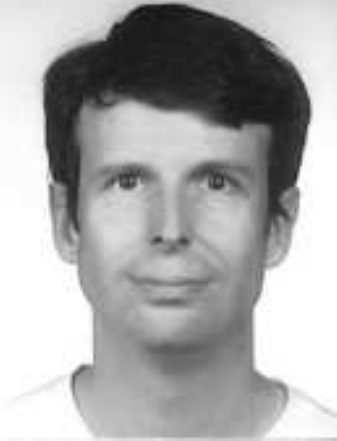}}]{Stephan Hartmann}
received the Dipl.-Ing. (M.Sc.) in Electrical Engineering from Technische Universität Dresden, Germany, in 2007. He is
currently working as research associate with the Chair of Highly-Parallel VLSI-Systems and Neuromorphic Circuits at Technische
Universität Dresden. His research interests include circuit design with special focus on FPGA.
\end{IEEEbiography}

\begin{IEEEbiography}[{\includegraphics[width=1in,height=1.25in,clip,keepaspectratio]{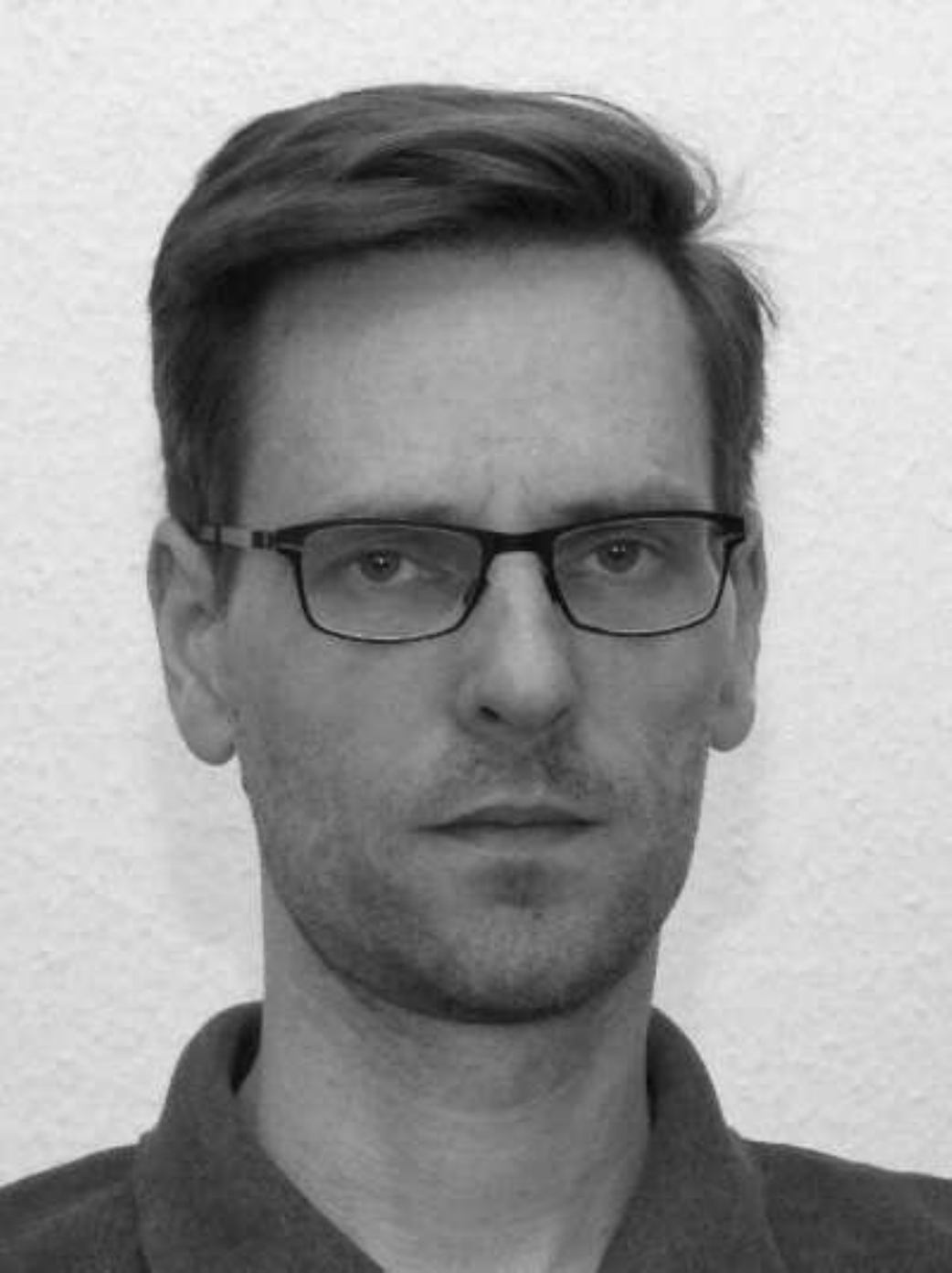}}]{Stefan Schiefer}
received the Dipl.-Ing. (M.Sc.) in Electrical Engineering from Technische Universität Dresden, Germany in 2008.
He is currently working as research associate with the Chair of Highly-Parallel VLSI-Systems and Neuromorphic Circuits at Technische Universität Dresden.
His research interests include full system signal and power integrity as well as regulator behavioural modelling including non-linearities.
\end{IEEEbiography}

\begin{IEEEbiography}[{\includegraphics[width=1in,height=1.25in,clip,keepaspectratio]{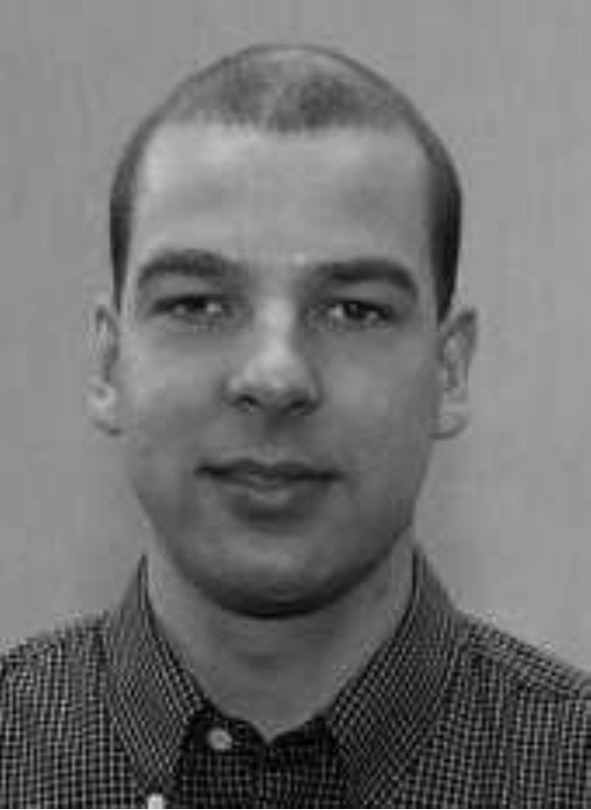}}]{Georg Ellguth}
received the Dipl.-Ing. (M.Sc.) in Electrical Engineering from Technische Universität Dresden, Germany in 2004. Since 2004, he has been a research assistant with the Chair of Highly-Parallel VLSI-Systems and Neuromorphic Circuits at Technische Universität Dresden. His research interests include low-power implementation techniques in multi-processor system-on-chip.
\end{IEEEbiography}

\begin{IEEEbiographynophoto}{Love Cederstroem (M'14)}
received his civ.ing.\ degree (M.Sc.) in Applied Physics and Electrical Engineering from Linköping University in 2009. In 2013 he joined the Chair of Highly-Parallel VLSI-Systems and Neuromorphic Circuits, Technische Universität Dresden, where he currently is a research assistant. His work focuses on system design and methodologies to bridge the chip-package-system boundaries, currently with emphasis on power delivery.
\end{IEEEbiographynophoto}

\begin{IEEEbiography}[{\includegraphics[width=1in,height=1.25in,clip,keepaspectratio]{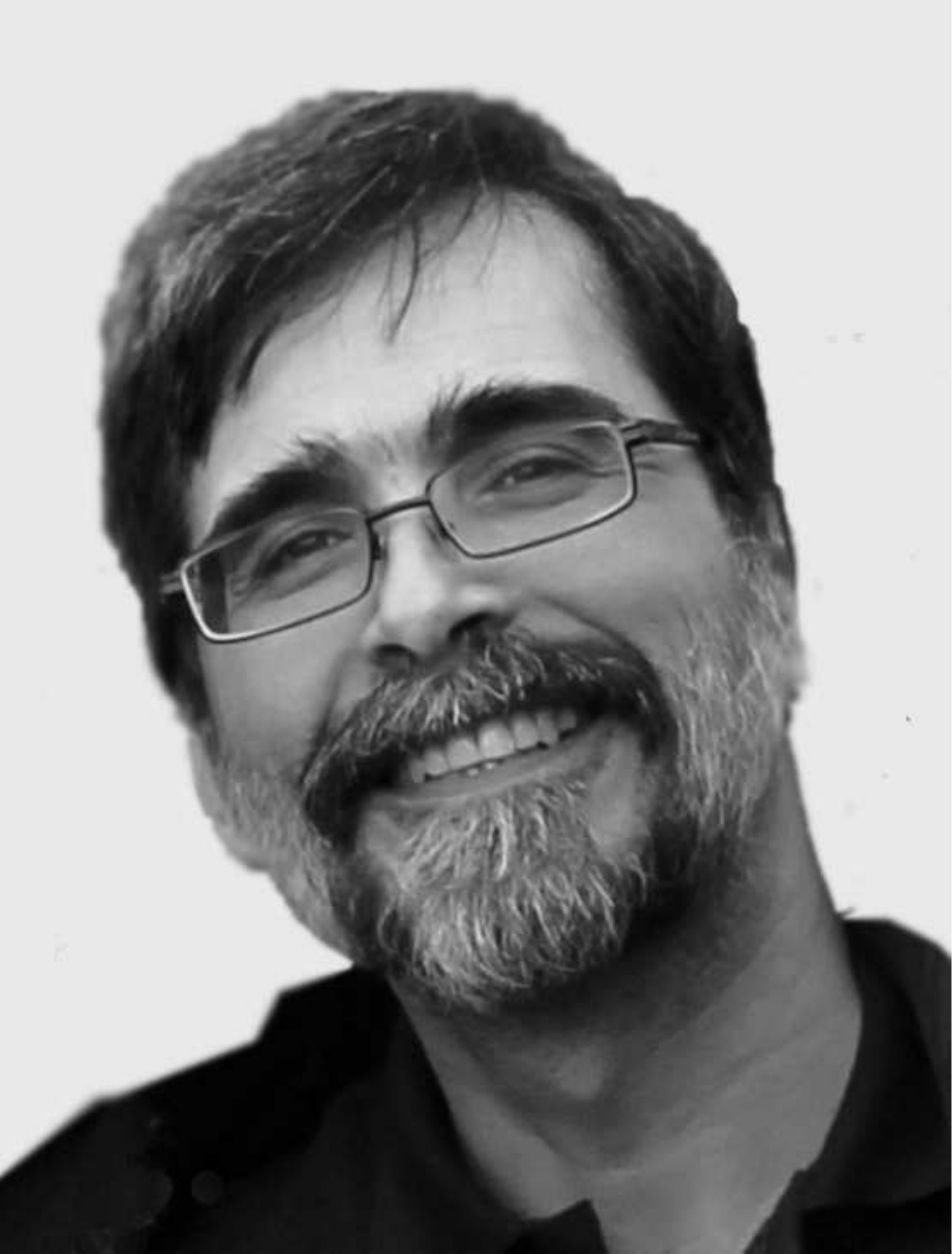}}]{Luis A Plana}
(M'97-SM'07) received the Ingeniero Electrónico (Cum Laude) degree from Universidad Simón Bolívar, Venezuela, and the PhD degree in computer science from Columbia University, USA.\@ He was with Universidad Politécnica, Venezuela, for over 20 years, where he was Professor of Electronic Engineering. Currently, he is a Research Fellow in the School of Computer Science, University of Manchester, UK.
\end{IEEEbiography}

\begin{IEEEbiography}[{\includegraphics[width=1in,height=1.25in,clip,keepaspectratio]{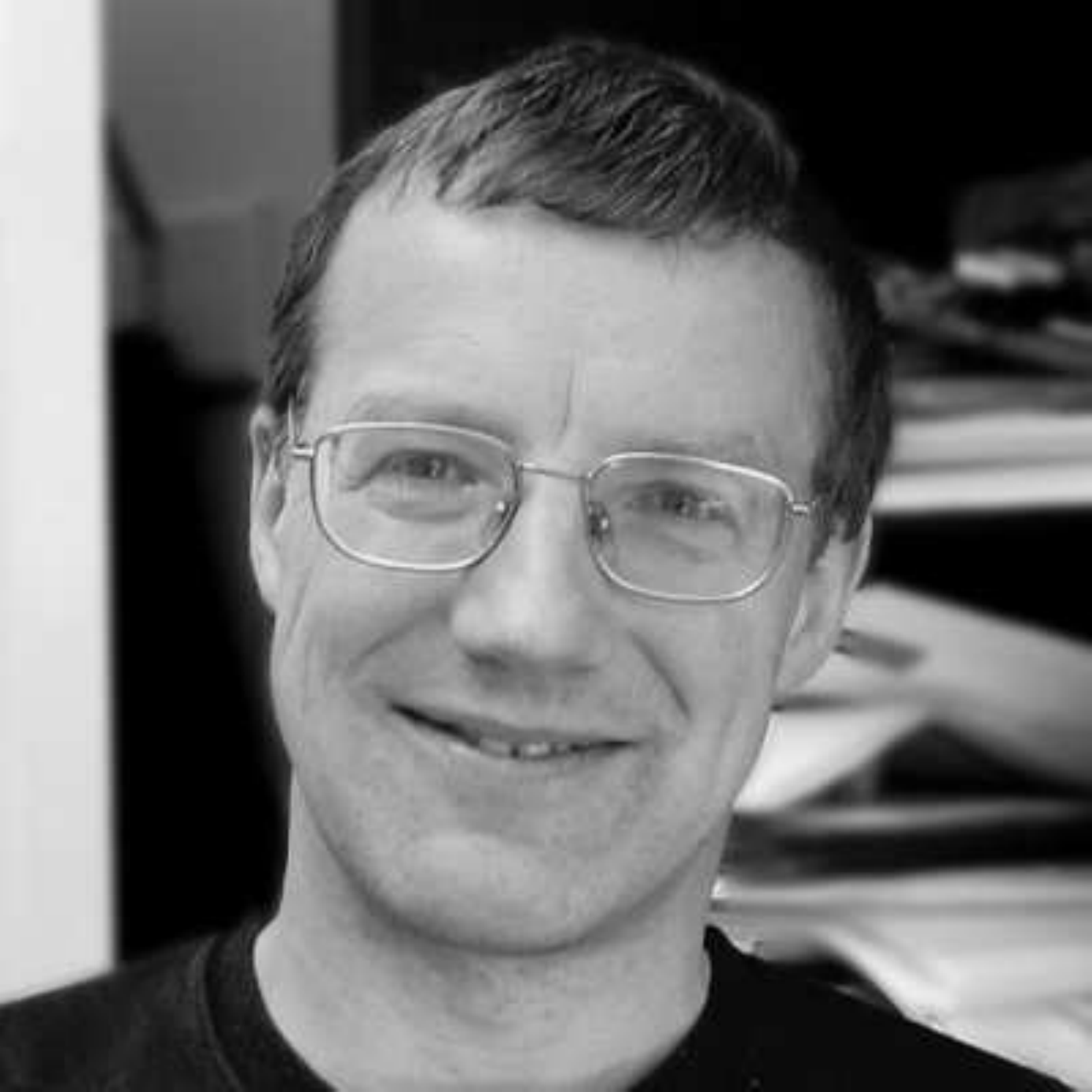}}]{Jim Garside}
received the Ph.D. degree in computer science from The University of Manchester, Manchester, U.K., in 1987, for work in signal processing architecture.  Post-doctoral work on parallel processing systems based on Inmos Transputers was followed by a spell in industry writing air traffic control software.  Returning to academia gave an opportunity for integrated circuit design work, dominated by design and construction work on asynchronous microprocessors in the 1990s. He has been involved with dynamic hardware compilation, GALS interconnection, and the development of the hardware and software of the SpiNNaker neural network simulator.
\end{IEEEbiography}

\begin{IEEEbiography}[{\includegraphics[width=1in,height=1.25in,clip,keepaspectratio]{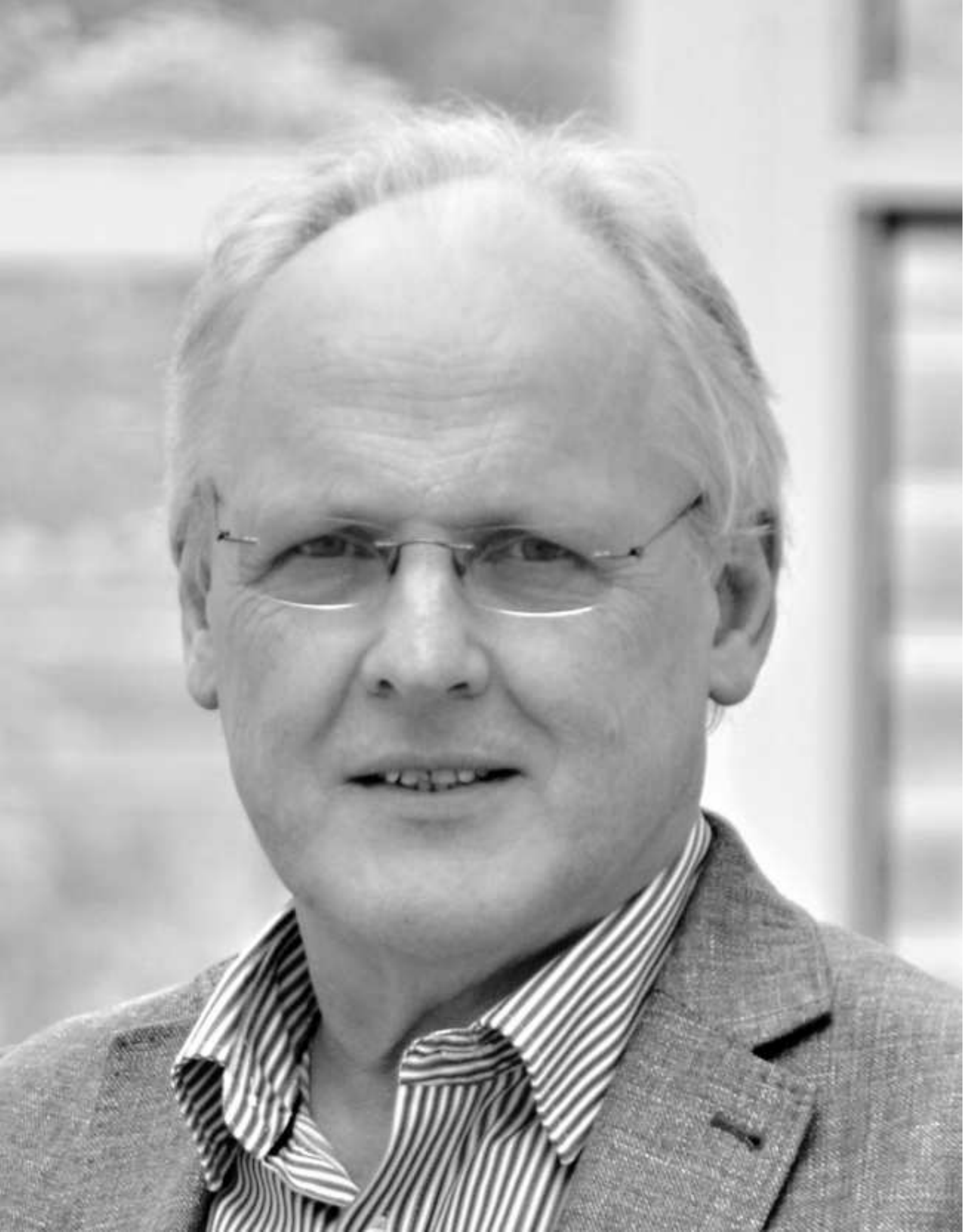}}]{Steve Furber}
CBE FRS FREng is ICL Professor of Computer Engineering in the School of Computer Science at the University of Manchester, UK. After completing a BA in mathematics and a PhD in aerodynamics at the University of Cambridge, UK, he spent the 1980s at Acorn Computers, where he was a principal designer of the BBC Microcomputer and the ARM 32-bit RISC microprocessor. Over 120 billion variants of the ARM processor have since been manufactured, powering much of the world's mobile and embedded computing. He moved to the ICL Chair at Manchester in 1990 where he leads research into asynchronous and low-power systems and, more recently, neural systems engineering, where the SpiNNaker project is delivering a computer incorporating a million ARM processors optimised for brain modelling applications.
\end{IEEEbiography}

\begin{IEEEbiography}[{\includegraphics[width=1in,height=1.25in,clip,keepaspectratio]{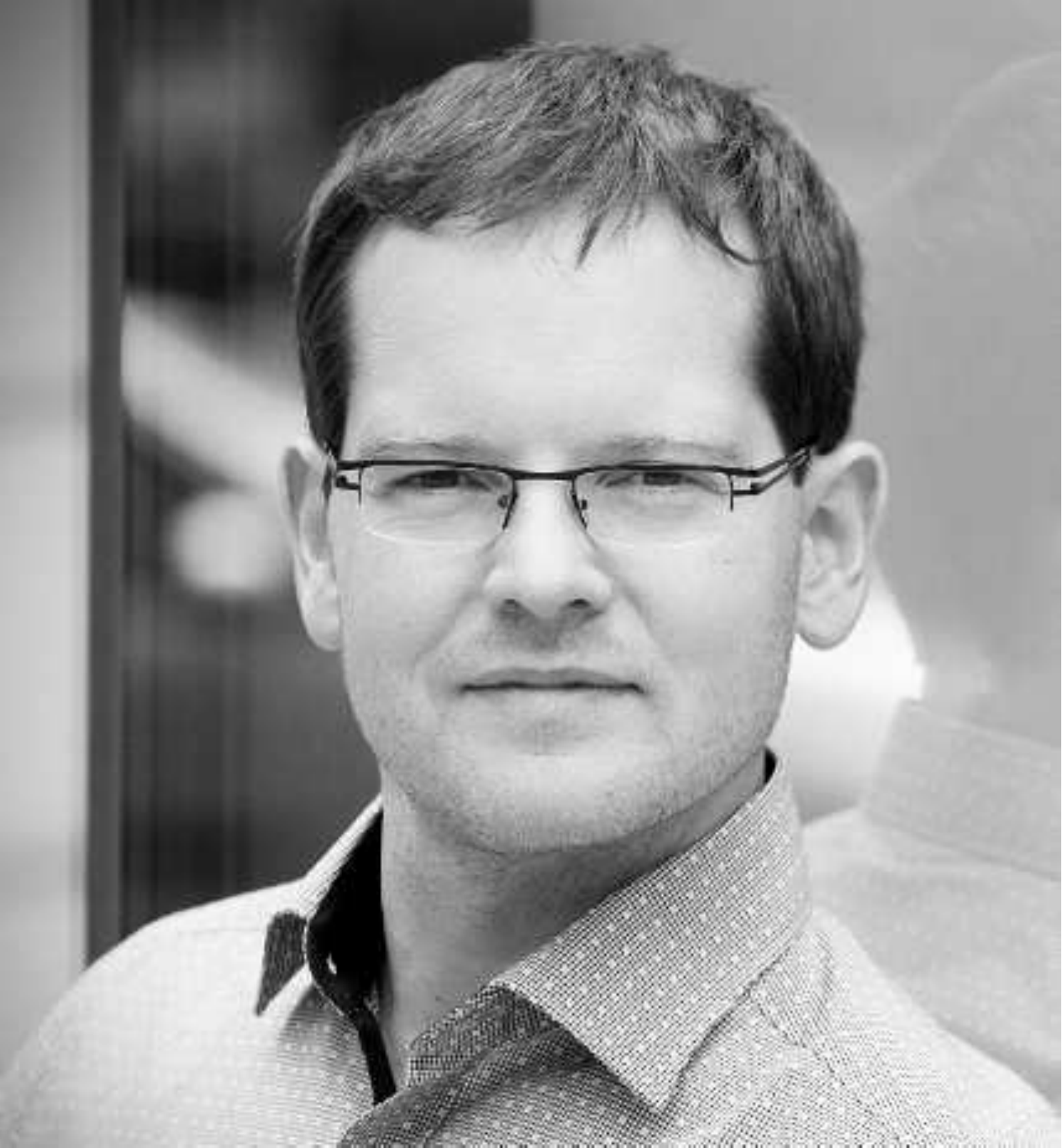}}]{Christian Mayr}
is a Professor of Electrical Engineering at TU Dresden. He received the Dipl.-Ing. (M.Sc.) in Electrical Engineering in 2003, his PhD in 2008 and Habilitation in 2012, all three from Technische Universität Dresden, Germany.
From 2003 to 2013, he has been with Technische Universität Dresden, with a secondment to Infineon (2004-2006). From 2013 to 2015, he did a Postdoc at the Institute of Neuroinformatics, University of Zurich and ETH Zurich, Switzerland. Since 2015, he is head of the Chair of Highly-Parallel VLSI-Systems and Neuromorphic Circuits at Technische Universität Dresden. His research interests include bio-inspired circuits, brain-machine interfaces, AD converters and general mixed-signal VLSI-design. He is author/co-author of over 80 publications and holds 4 patents. He has acted as editor/reviewer for various IEEE and Elsevier journals. His work has received several awards.
\end{IEEEbiography}